\documentclass[lettersize,journal]{IEEEtran}
\usepackage{graphicx} 
\usepackage{lipsum} 
\usepackage{subcaption}
\usepackage{amsmath,amsfonts}
\usepackage{algorithmic}
\usepackage{algorithm}
\usepackage{array}
\usepackage[caption=false,font=normalsize,labelfont=sf,textfont=sf]{subfig}
\usepackage{textcomp}
\usepackage{stfloats}
\usepackage{url}
\usepackage{verbatim}
\usepackage{graphicx}
\usepackage{cite}
\usepackage{caption}
\usepackage{booktabs}
\usepackage{amssymb} 
\usepackage{color}
\usepackage{colortbl}
\usepackage[table]{xcolor} 

\begin{document}
	\bibliographystyle{unsrt}
	
	\title{FusionTrack: End-to-End Multi-Object Tracking in Arbitrary Multi-View Environment}
	\author{Xiaohe Li, Pengfei Li, Zide Fan, ~\IEEEmembership{Member,~IEEE}, Ying Geng, Fangli Mou, ~\IEEEmembership{Member,~IEEE}, Haohua Wu,\\ Yunping Ge
		\thanks{Xiaohe Li, Pengfei Li, Zide Fan, Ying Geng, Fangli Mou, Haohua Wu, Yunping Ge are with Aerospace Information Research Institute, Chinese Academy of Sciences (e-mail: lixiaohe@aircas.ac.cn, lipengfei23@mails.ucas.ac.cn, fanzd@aircas.ac.cn, geyp@aircas.ac.cn).}
		\thanks{\textit{(corresponding author: Zide Fan)} }
		\thanks{\textit{(Xiaohe Li and Pengfei Li contribute equally to the article)}}
		\thanks{This work has been submitted to the IEEE for possible publication. Copyright may be transferred without notice, after which this version may no longer be accessible.}
	}
	\markboth{Journal of \LaTeX\ Class Files,~Vol.~14, No.~8, August~2021}%
	{Shell \MakeLowercase{\textit{et al.}}: A Sample Article Using IEEEtran.cls for IEEE Journals}
	
	
	\maketitle
	
	\begin{abstract}
		Multi-view multi-object tracking (MVMOT) has found widespread applications in intelligent transportation, surveillance systems, and urban management. However, existing studies rarely address genuinely free-viewpoint MVMOT systems, which could significantly enhance the flexibility and scalability of cooperative tracking systems. 
		To bridge this gap, we first construct the Multi-Drone Multi-Object Tracking (MDMOT) dataset, captured by mobile drone swarms across diverse real-world scenarios, initially establishing the first benchmark for multi-object tracking in arbitrary multi-view environment. Building upon this foundation, we propose \textbf{FusionTrack}, an end-to-end framework that reasonably integrates tracking and re-identification to leverage multi-view information for robust trajectory association.
		Extensive experiments on our MDMOT and other benchmark datasets demonstrate that FusionTrack achieves state-of-the-art performance in both single-view and multi-view tracking.
	\end{abstract}
	
	\begin{figure*}[h]
		\centering
		\includegraphics[width=7in]{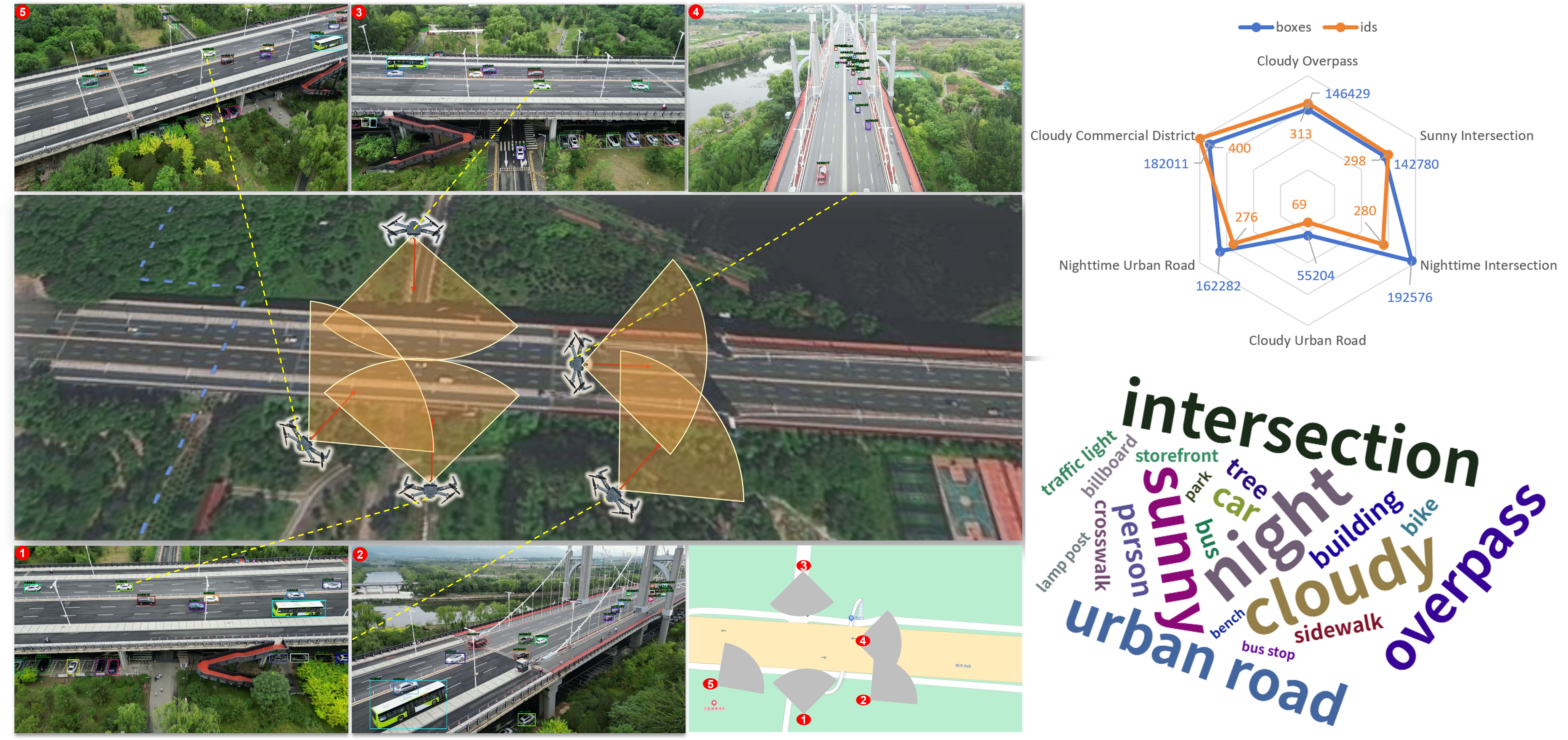}
		\caption{Examples of the MDMOT dataset. The left panel illustrates the spatial distribution of drones at a specific moment within the overpass scenario, along with the corresponding multi-view imagery they captured, clearly highlighting both overlapping and non-overlapping regions. The top-right quadrant presents statistical analysis of bounding box distributions and identity frequencies across six representative training scenes. The bottom-right section displays a word cloud visualizing prominent feature terms in the dataset.}
		\label{view}
	\end{figure*}
	\section{Introduction}
	
	
	

	\IEEEPARstart{M}ulti-object tracking (MOT) focuses on accurately localizing and consistently identifying multiple targets within video sequences\cite{bashar2022multiple}, and has been widely adopted in applications such as video surveillance\cite{yang2014robust,fan2024gmt}, intelligent transportation systems\cite{zhang2024video,sun2022drone}, behavior analysis\cite{ma2019bayesian,sun2024multiple}, and scene understanding\cite{tang2019cityflow}. Nevertheless, the limited detection range from single-view observations often fall short in meeting the demands of large-scale, complex urban monitoring. To address this, multi-view multi-object tracking (MVMOT) leveraging multiple drone cameras has garnered increasing attention\cite{liu2023robust}. At its core, multi-view tracking entails the integration of tracklet estimation and identity association across views, allowing for more continuous and robust tracking by harnessing broader spatial coverage and collaborative sensing. However, it also presents new challenges, including view-dependent random occlusion, variations in coverage range, and appearance discrepancies caused by differing perspectives, all of which complicate reliable cross-view target association.  
	
	Based on the patterns of data acquisition, MVMOT can be broadly divided into two categories: overlapping-view and non-overlapping-view\cite{amosa2023multi}. As illustrated in Fig. \ref{view}, Drones $\sharp$1, $\sharp$3, and $\sharp$5 share overlapping fields of view, while Drones $\sharp$2 and $\sharp$4 form another overlapping group. Notably, there is no overlap between these two groups, resulting in fully disjoint viewing areas. Currently, the majority of existing research focuses on overlapping-view settings\cite{chen2016integrating}, including pedestrian tracking datasets such as EPFL\cite{fleuret2007multicamera}, CAMPUS\cite{xu2016multi}, MvMHAT\cite{gan2021self}, and MMP-Tracking\cite{han2023mmptrack}, captured by ground-based cameras, as well as drone-based multi-view datasets like DIVOTrack\cite{hao2024divotrack}, MDMT\cite{liu2023robust}, and VisionTrack\cite{fan2024gmt}. Meanwhile, some efforts have leveraged the CityFlow\cite{tang2019cityflow} dataset from the AI City Challenge, which comprises multi-view tracking data across several city blocks with completely non-overlapping fields of view. Building upon this, the HST\cite{zhang2024video} dataset was developed, capturing highway tunnel traffic through cameras deployed at multiple positions. Beyond conventional urban scenarios, some datasets extend to more diverse settings. For example, BuckTales\cite{naik2024bucktales} collects wildlife footage using multiple drones. MITracker\cite{xu2025mitracker} introduces a multi-view tracking dataset involving 27 object categories, though its scenes are primarily constrained to controlled experimental environments. Despite these developments, the previous multi-view tracking datasets lack research on complex situations, especially in terms of datasets with unconstrained viewpoints (including overlapping-view and non-overlapping-view) that better reflect real-world deployment. To overcome these challenges, we propose a new formulation of MOT under arbitrary numbers of views, and present a novel Multi-Drone Multi-Object Tracking (MDMOT) dataset constructed using a fleet of drones. This dataset incorporates both overlapping and non-overlapping views, spanning a wide range of complex environments and dense target distributions. Compared with existing datasets, ours introduces additional challenges: (1) Targets can appear or disappear from any location, unconstrained by fixed-view assumptions. (2) High-speed moving targets are prone to motion blur, degrading recognition reliability. (3) The top-down perspective and elevated flight altitude result in small target sizes, increasing detection difficulty. (4) Dynamic changing viewpoints complicate cross-view association due to greater inter-view variability.
	
	Current research in MVMOT predominantly adopts a two-stage training framework following a track-and-associate paradigm. In the first stage, single-view MOT methods such as SORT\cite{bewley2016simple}, DeepSORT\cite{wojke2017simple}, and ByteTrack\cite{zhang2022bytetrack} are used to generate local trajectories. These trajectories are then associated across views using various strategies, including: (1) Re-identification(ReID)-based association, where visual similarity between trajectories is computed using ReID features. While widely used, this approach is highly dependent on the accuracy of the initial single-view tracking (SVT) results. (2) Image matching-based association, which is suitable for overlapping camera views and employs keypoint-based algorithms such as SIFT\cite{low2004distinctive} to estimate homography matrices for cross-view matching. However, this method requires overlapping fields of view and is computationally intensive. (3) Non-negative matrix factorization (NMF)\cite{he2020multi}, which infers associations directly by factorizing similarity matrices derived from tracklet data. While two-stage frameworks are favored for their modularity and interpretability, the decoupled nature of their components means that performance limitations in any stage can hinder overall system effectiveness. To address this, some end-to-end approaches have been proposed \cite{nguyen2024lammon}, which unify tracking and association into a single training pipeline by graph-based representations. These unified models offer improved global context modeling capabilities compared to their two-stage counterparts. Nevertheless, their high computational cost remains a significant barrier to real-time deployment.
	
	Motivated by the recent advances of Transformer-based architectures in both 2D and 3D object detection and tracking, we propose FusionTrack—a novel end-to-end framework for MOT across arbitrary numbers of views. Unlike traditional two-stage pipelines, FusionTrack eliminates the need for multi-step training and instead leverages the strong global modeling capability of Transformer to jointly optimize SVT and cross-view ReID within a unified architecture. To enhance feature representation, we introduce a feature aggregation module that dynamically integrates features of the same identity observed across different timestamps and views, thereby enriching the current query embeddings. In addition, we deploy a tracklet memory pool to maintain temporal continuity and address memory-based ReID and target reentry across views. For identity association, we design an optimization strategy based on optimal transport distance matrices, facilitating robust global identity matching.
	During inference, we implement an efficient post-processing pipeline that combines view-aware masking strategies, a neighborhood-based filtering mechanism, and view-guided hierarchical clustering, achieving accurate and scalable cross-view target association.
	
	To summarize, our key contributions are as follows:
	\begin{itemize}
		\item We introduce the problem of MOT under arbitrary numbers of views and construct a new multi-view dataset named \textbf{MDMOT} captured by a fleet of drones. The dataset includes both overlapping and non-overlapping camera configurations, and spans a diverse range of real-world environments.
		\item We develop \textbf{FusionTrack}, an attention-based end-to-end framework for multi-view tracking that jointly optimizes tracking and ReID. The framework incorporates a novel Object Update Module (OUM), a Tracklet Memory Pool (TMP), and a global identity optimization strategy based on optimal transport distance matrices.
		\item We design an efficient post-processing pipeline that enforces intra-view exclusivity and association uniqueness through masking rules, and introduces a Neighbor Filtering Mechanism (NFM) to suppress erroneous identity matches.
		\item  Experimental results on multiple benchmark datasets show that our method achieves state-of-the-art results in both single-view and multi-view tracking.
	\end{itemize}

	\section{RELATED WORK}
	
	\subsection{Object detection and tracking}
	
	
	Object detection serves as a cornerstone task in computer vision, focusing on the localization and classification of objects within images\cite{li2024open}. With the surge of deep learning, object detection techniques have witnessed significant advancements and have reached a high level of maturity. Existing approaches are commonly divided into two categories: two-stage and one-stage detectors. Two-stage frameworks, exemplified by Faster R-CNN\cite{girshick2015fast} and Cascade R-CNN\cite{cai2018cascade}, first generate region proposals followed by precise localization and classification, offering high accuracy at the expense of increased computational cost. In contrast, one-stage detectors such as YOLO\cite{jiang2022review} and SSD\cite{liu2016ssd} predict bounding boxes and class probabilities in a single forward pass, achieving a better trade-off between speed and accuracy.  Subsequently, a series of Transformer-based end-to-end object detection methods were proposed, demonstrating strong performance by effectively leveraging their long-range dependency modeling capabilities\cite{yuan2023ctif}.
	
	As a downstream task of object detection, object tracking aims to continuously locate and associate targets in video sequences. Currently, mainstream tracking methods mainly follow two paradigms: tracking-by-detection and end-to-end approaches. In general, the former tends to achieve higher accuracy but is heavily reliant on detection results, whereas the latter avoids information loss caused by stepwise optimization and offers higher efficiency.
	The tracking-by-detection paradigm first performs object detection on each frame, followed by target association using methods such as Kalman filtering and the Hungarian algorithm. For example, SORT\cite{bewley2016simple} combines Kalman filtering with a simple IoU-based association strategy to achieve efficient tracking. DeepSORT\cite{wojke2017simple} further incorporates deep appearance features to improve matching accuracy. And ByteTrack\cite{zhang2022bytetrack} addresses occlusion and missing detections by performing two-stage association between low- and high-confidence detections.
	In contrast, end-to-end methods integrate detection and association within a unified framework, reducing the reliance on traditional matching algorithms. For instance, Tracktor\cite{sridhar2019tracktor} propagates detections by regressing bounding boxes from previous frames. JDE\cite{wang2020towards} and FairMOT\cite{zhang2021fairmot} enhance appearance embedding and association accuracy by jointly optimizing detection and ReID tasks.
	In recent years, Transformer structures have been widely adopted in computer vision, leading to the development of Transformer-based end-to-end tracking frameworks such as TransTrack\cite{sun2020transtrack}, TrackFormer\cite{meinhardt2022trackformer}, MOTR\cite{zeng2022motr}, and MeMOTR\cite{gao2023memotr}. These methods formulate targets as queries and leverage the long-range dependency modeling capability of Transformer to unify detection and association, significantly improving both tracking performance and runtime efficiency.In addition, several recent studies, including MambaVT\cite{lai2024mambavt}, have explored the fusion of RGB and infrared modalities to enhance object tracking performance.
	\subsection{Object ReID}
	
	
	
	ReID serves as a crucial component of MVMOT, aiming to retrieve the same identity across varying camera perspectives or scene contexts\cite{zhang2025mask}. The task is inherently challenging due to factors such as viewpoint shifts, illumination changes, occlusions, and pose variations.
	Existing ReID methods can be broadly divided into two categories: global feature-based and local feature-based approaches. Global feature methods represent the target holistically, extracting a single feature vector using convolutional neural networks. These approaches are computationally efficient and perform well when the target is fully visible. However, they are sensitive to occlusions and drastic pose changes, which often compromise their effectiveness.
	To address these issues, local feature-based methods segment the target into multiple regions, independently extract local descriptors, and subsequently fuse them to capture fine-grained semantic cues. Notable examples include PCB\cite{sun2018beyond}, MGN\cite{wang2018learning}, SAN\cite{jin2020semantics}, and AGW\cite{ye2021deep}. PCB\cite{sun2018beyond} applies a uniform horizontal slicing strategy to divide the image into six stripes and learns part-level features. MGN\cite{wang2018learning} employs a multi-branch network to capture representations at varying granularities. SAN\cite{jin2020semantics} leverages semantic attention to focus on discriminative regions. And AGW\cite{ye2021deep} incorporates attention mechanisms to integrate global context with local details. These methods have demonstrated enhanced robustness under challenging conditions such as occlusion and pose variation, owing to their ability to capture complementary information from different parts of the target.
	
	By dividing images into sequences of patches and modeling long-range token dependencies, Transformer structures effectively capture semantic relationships across spatial regions.
	TransReID\cite{he2021transreid} pioneered the application of pure Transformer architectures to the ReID task. It partitions input images into patch tokens for localized representation learning and incorporates a CLS token for global identity prediction. The model is jointly trained using identity classification and triplet losses, yielding state-of-the-art results.
	Following this work, numerous Transformer-based ReID methods have been proposed, further advancing the performance frontier. Moreover, hybrid models that integrate CNNs and Transformers have gained traction, aiming to exploit complementary strengths across spatial and temporal domains. For instance, HAT\cite{zhang2021hat} introduces a hybrid framework that leverages Transformers to aggregate multi-scale CNN features from a global perspective, thereby improving the robustness and expressiveness of learned representations.
	
	\subsection{Multi-view multi-object tracking}

	
	Similar to SVT, MVMOT can be broadly categorized into two-stage and end-to-end approaches. Two-stage frameworks typically decompose the task into SVT and inter-view tracking (IVT). The latter aims to associate identical targets across different camera views to construct globally consistent trajectories.
	A number of strategies have been proposed for cross-view tracklet matching\cite{you2020multi}, including hierarchical matching\cite{li2019spatio}, data association graphs\cite{chen2016equalized}, and camera link models\cite{hofmann2013hypergraphs}, all aiming to enhance trajectory association across views.
	Despite these efforts, cross-view ReID remains challenging due to variations in viewpoint, lighting, and occlusions. To mitigate these issues, recent MVMOT studies have focused on enhancing feature representation and association strategies.
	For instance, DyGLIP\cite{quach2021dyglip} leverages a dynamic graph neural network with attention to improve association accuracy. TRACTA\cite{he2020multi} formulates MVMOT as a constrained tracklet-to-target matching problem and solves it using non-negative matrix factorization. MvMHAT\cite{gan2021self} introduces a self-supervised framework that performs cross-view association through pairwise and triplet-based similarity learning. CityTrack\cite{yang2022box} adopts a two-stage pipeline and integrates traffic rule constraints for multi-camera vehicle tracking. To address the difficulty of small-object tracking, MIA-NET\cite{liu2023robust} proposes a keypoint alignment strategy that promotes cross-view consistency in multi-view settings.
	
	While the methods discussed above have shown effectiveness in certain scenarios, they often suffer from error propagation between the SVT and IVT stages, which hampers the utilization of rich spatiotemporal correlations across views. To overcome this limitation, growing efforts have shifted toward unified end-to-end frameworks.
	Several recent works have leveraged graph neural networks to directly model inter-object relationships and generate globally consistent trajectories. For example, Chen et al.\cite{chen2016equalized} proposed a global graph network that jointly optimizes SVT and IVT by balancing similarity metrics; Liu et al.\cite{liu2017multi} adopted a generalized maximum clique formulation to construct a unified graph-based model; and CrossMOT\cite{hao2024divotrack} enhances ReID discriminability by jointly optimizing intra- and inter-view embedding spaces during tracking, leading to improved performance in pedestrian MVMOT tasks.
	Although these methods have pushed the field forward, they commonly treat tracking and ReID as decoupled components, failing to fully leverage their potential mutual reinforcement. Inspired by the growing success of Transformer architectures in both MOT and ReID, we propose a novel end-to-end Transformer-based framework for MVMOT that systematically addresses these issues. To the best of our knowledge, this is the first fully Transformer-based end-to-end MVMOT model, and extensive experiments on multiple datasets validate its effectiveness.

	\section{MDMOT DATASET}
	To advance research in multi-view object tracking, we present  \textbf{M}ulti-\textbf{D}rones \textbf{M}ulti-\textbf{O}bject \textbf{T}racking (\textbf{MDMOT})—a newly curated benchmark dataset captured through coordinated multi-drone operations. To the best of our knowledge, MDMOT is the first dataset that supports arbitrary and unconstrained viewpoints in multi-drone multi-object tracking scenarios. In what follows, we elaborate on the data acquisition process, annotation protocol, dataset statistics, and its defining characteristics. 
	
	\subsection{Data Collection and Annotation}
	\subsubsection{Data Collection}
	The MDMOT dataset was constructed using a coordinated fleet of 5 DJI Mini Pro 3 drones, capturing a total of 20 synchronized multi-view video sequences at a resolution of $1920\times1080$. The videos span 7 distinct real-world urban environments, including overpasses, intersections, commercial zones, and city streets. Each scene comprises 3 to 4 video segments, with 3 to 5 drones recording simultaneously per clip, resulting in more than 120,000 annotated frames. Fig. \ref{view} shows an example of the spatial deployment of drone viewpoints and the corresponding multi-view images for an overpass scenario.
	
\begin{figure}[!h]
		\centering
		\includegraphics[width=3.2in]{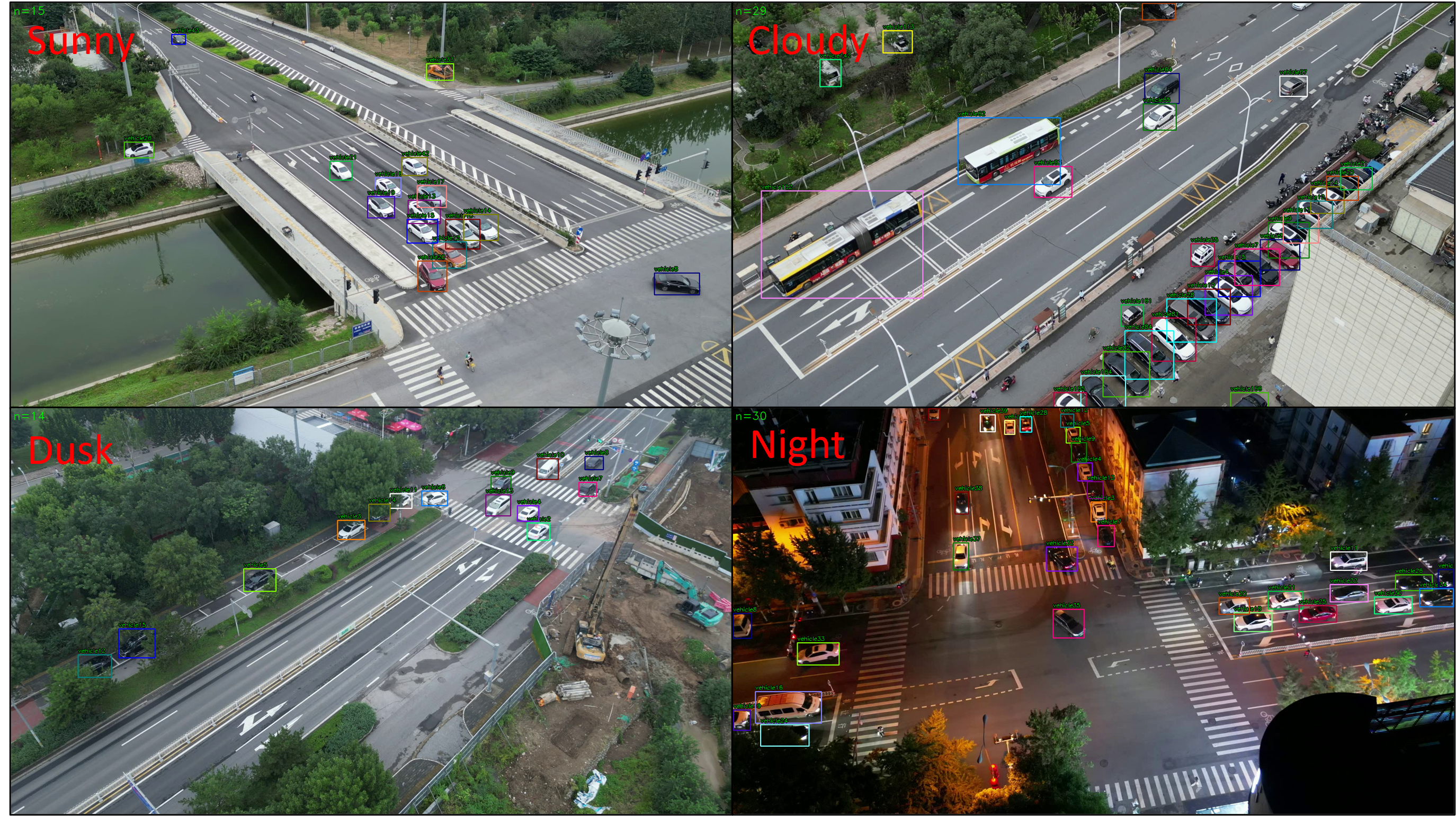}
		\caption{Image examples under varying weather conditions. From top-left to bottom-right: sunny, cloudy, dusk, and nighttime.}
		\label{statistic1}
	\end{figure}

	To reflect the real-world challenges of MVMOT, we select urban traffic environments characterized by dense target presence and complex dynamics, including occlusion, high-speed motion, and interweaving trajectories. 
	Data was captured by drones manually operated under professional guidance, with flight altitudes ranging from 30 to 100 meters. A combination of top-down and oblique camera angles was employed to ensure diverse viewpoint coverage.
	To further enhance the dataset’s robustness and adaptability, we hold a range of weather and illumination conditions, such as sunny, cloudy, dusk, and nighttime settings, as illustrated in Fig. \ref{statistic1}. These variations not only increase the dataset’s diversity but also bring it more adaptable to real-world deployment scenarios.

	Regarding viewpoint configuration, the dataset includes both partially overlapping views, which facilitate cross-camera identity matching, and completely non-overlapping views, which pose more challenging conditions for performance evaluation. This dual-design setup makes the MDMOT dataset a versatile and comprehensive benchmark for MVMOT, suitable for advancing research in MOT, cross-view ReID and collaborative drone-based surveillance systems.  
	
\begin{table*}[!t]
	\caption{COMPARISON OF MDMOT DATASET AND EXISTING DATASETS. }
	\centering
	\begin{tabular}{ccccccccc}
		\toprule
		Dataset &Source&Resolustion&Scenes&Views&Frames&Boxes& Moving Cameras& View distribution\\
		\midrule
		CityFlow\cite{tang2019cityflow}&Monitor&1280$\times$960&1&46&117K&230K&$\times$ & overlap\&non-overlap  \\
		EPFL\cite{fleuret2007multicamera}&Monitor&360$\times$288&5&2-4&97K&625k&$\times$ & overlap  \\
		CAMPUS\cite{xu2016multi}&Monitor&1920$\times$1080&4&4&83K&490K&$\times$ &  overlap\&non-overlap  \\
		MvMHAT\cite{gan2021self}&Monitor&1920$\times$1080&1&3-4&31K&208K&\checkmark &  overlap\&non-overlap  \\
		WILDTRACK\cite{chavdarova2017wildtrack}&Monitor&1920$\times$1080&1&7&3K&40K&$\times$ & overlap  \\
		DIVOTrack\cite{hao2024divotrack}&Mobile,Drone&3640$\times$2048,1920$\times$1080&10&3&81K&830K&\checkmark & overlap  \\
		VisionTrack\cite{fan2024gmt}&Drone&1920$\times$1080&15&2&116K&1176K&\checkmark& overlap \\
		MDMT\cite{liu2023robust}&Drone&1920$\times$1080&6&2&40K&220K&\checkmark& overlap \\
		\midrule
		\rowcolor{gray!20}
		MDMOT&Drone&1920$\times$1080&6&3-5&122K&2481K& \checkmark& overlap\&non-overlap   \\
		\bottomrule
		\label{dataset}
	\end{tabular}
\end{table*}
	
	\subsubsection{Data annotation}
	
	To ensure temporal consistency across multiple views, we first execute timestamp-based alignment and trimming of the video clips prior to annotation. Leveraging timestamps provided by the drones, we align the start and end times across different camera views, ensuring that all sequences are temporally synchronized.
	To further enhance the quality and applicability of the dataset, a manual screening process was conducted to filter out low-quality video segments. Footage with insufficient targets, poor visual quality, inadequate lighting, or severe motion blur was excluded. Only high-quality clips are retained for subsequent annotation and evaluation.
	In the final dataset, each processed video segment contains between 900 and 3000 frames, with the frame count synchronized across all views within the same segment, facilitating efficient multi-view annotation and association tasks.
	
	To ensure both efficiency and precision in annotation, we adopt a semi-automatic labeling pipeline that integrates automatic tracking with human verification. We first perform inference using a ByteTrack\cite{zhang2022bytetrack} model pretrained on the VisDrone dataset \cite{wen2019visdrone} to obtain initial tracking predictions for each video segment.
	Subsequently, the predictions are refined using the DarkLabel annotation tool, where annotators manually address key issues such as identity switches, inaccurate bounding box dimensions, and both false positives and missed detections.
	After completing per-view annotations, we manually associate corresponding targets across different views to guarantee identity consistency in the multi-view setting.
	
	The finalized annotation format follows the structure: $⟨frame, id, ncx, ncy, nw, nh⟩$, where $frame$ corresponds to the frame number, $id$ denotes the unique identity of the target, and $ncx$, $ncy$ indicate the normalized x and y coordinates of the bounding box center. $nw$ and $nh$ refer to the normalized width and height of the bounding box, respectively. This compact format is optimized for training modern tracking models and facilitates cross-view association.
	
	\subsection{Statistics and Splits}
	
	The mobility and viewpoint flexibility of drones introduce high spatial uncertainty in target appearances—objects may emerge in virtually any region of the image. As shown in Fig. 3(a), the heatmap generated from all annotated targets in MDMOT reveals a broad and relatively uniform distribution of targets across the image plane.
	In contrast to datasets collected from static surveillance cameras, MDMOT lacks any fixed spatial priors for target placement. This inherent variability elevates the dataset’s complexity and better reflects the challenges of real-world drone-based tracking scenarios.
	
	Beyond spatial randomness, large variations in target sizes pose an additional challenge in MDMOT. As illustrated in Fig. 3(b), the bounding box size distribution reveals substantial scale differences among targets, primarily caused by varying distances between the drones and objects. This leads to a highly imbalanced distribution of object sizes across the dataset.
	Such scale variability significantly complicates the task of accurate detection and consistent identity tracking, especially under real-world conditions.

	\begin{figure}[h] 
		\centering
		\subcaptionbox{}[.45\linewidth][c]{%
			\includegraphics[width=1.1\linewidth]{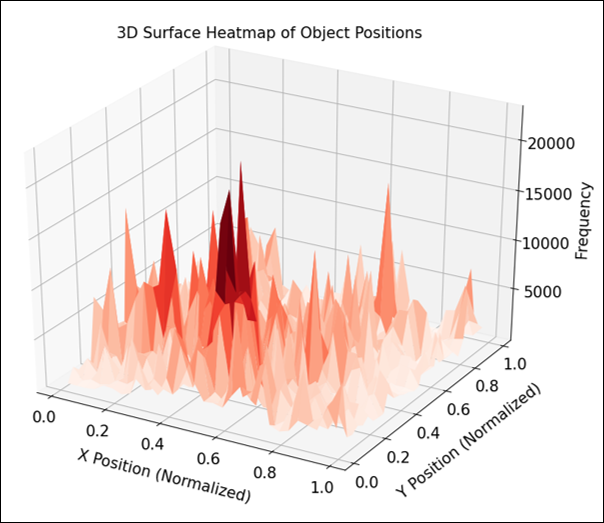}
		}\qquad
		\subcaptionbox{}[.45\linewidth][c]{%
			\includegraphics[width=1.1\linewidth]{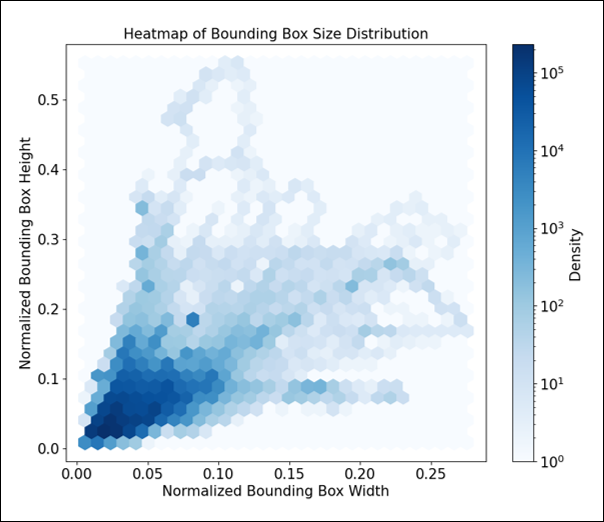}
		}
		\caption{(a) Heatmap of object location distribution, where the x and y axes represent normalized image coordinates, and the z-axis indicates the frequency of object occurrence. (b) Heatmap of box size distribution, with the x and y axes indicating the relative width and height of boxes.}
		\label{statistic2}
	\end{figure}
	
\begin{figure}[t]
		\centering
		\includegraphics[width=3.4in]{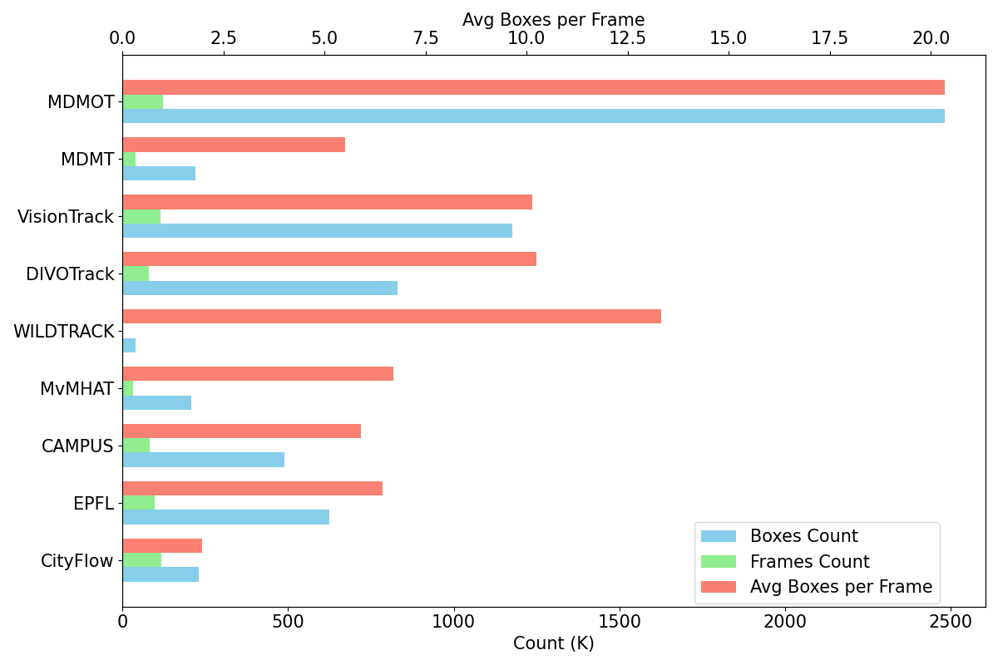}
		\caption{Comparison of dataset statistics, with blue indicating the total number of bounding boxes, green representing the total number of frames, and red denoting the average number of boxes per frame.}
		\label{statistic3}
	\end{figure}

	To enable robust training and evaluation of tracking algorithms in dense and dynamic environments, MDMOT offers a large-scale, high-density collection of video data. The dataset comprises approximately 122,000 frames, with over 24,810,000 annotated bounding boxes, resulting in an average of more than 20 targets per frame.
	As illustrated in Table I and Fig. \ref{statistic3}, MDMOT outperforms existing datasets in key aspects such as total frame volume, number of tracked instances, and target density per frame, making it a highly valuable resource for advancing research in MVMOT.
	To enhance the dataset’s generalization potential, MDMOT incorporates substantial diversity across both scene types and acquisition conditions. We curated four prototypical urban environments: overpasses, intersections, city streets, and commercial areas. In the intersection scenario, we further introduce four distinct weather conditions—sunny, cloudy,dusk and nighttime—to simulate a broader range of real-world challenges. For standardized evaluation, the dataset is partitioned into training, validation, and test splits with a frame ratio of approximately $8:1:1$.

	\subsection{Characteristics}
	Several multi-view tracking datasets have been introduced in prior work, including EPFL\cite{fleuret2007multicamera}, CAMPUS\cite{xu2016multi}, MvMHAT\cite{gan2021self}, and WILDTRACK\cite{chavdarova2017wildtrack} for pedestrian tracking, as well as CityFlow\cite{tang2019cityflow} for cross-view vehicle tracking. However, these datasets are predominantly constructed using fixed, statically deployed camera networks.
	Although such predefined camera layouts can improve target association accuracy, they inherently lack flexibility and struggle to adapt to complex and dynamic real-world environments.
	
	With the capability to dynamically adjust flight paths, camera angles, and coverage areas, drones have become a promising platform for large-scale dynamic scene understanding. Notably, DIVOTrack\cite{hao2024divotrack} pioneer the integration of heterogeneous sensing by combining aerial and ground-based views, while VisionTrack\cite{fan2024gmt} and MDMT\cite{liu2023robust} employ dual-drone setups to showcase the benefits of mobile and adaptive viewpoints in MOT.
	Despite these advances, current multi-drone datasets are still constrained by limited camera viewpoints, restricted coverage areas, and a general lack of realism and environmental complexity, hindering their applicability in more general scenarios.
	
	To overcome the limitations of prior datasets, we introduce MDMOT, a novel benchmark collected using a coordinated network of 3 to 5 drones. This setup enables the same target to be captured under diverse viewpoints, reflecting a broader range of appearance and motion patterns, while also introducing realistic complexities such as multi-object occlusions and sudden illumination changes.
	By offering greater viewpoint diversity, ensuring spatio-temporal consistency, and accommodating dynamic environmental conditions, MDMOT serves as a more representative and generalizable benchmark for evaluating algorithms. It provides the foundation for pushing multi-view tracking research closer to deployment in real-world, open-domain scenarios.

	\section{METHOD}
	\begin{figure*}[t]
		\centering
		\includegraphics[width=7.2in]{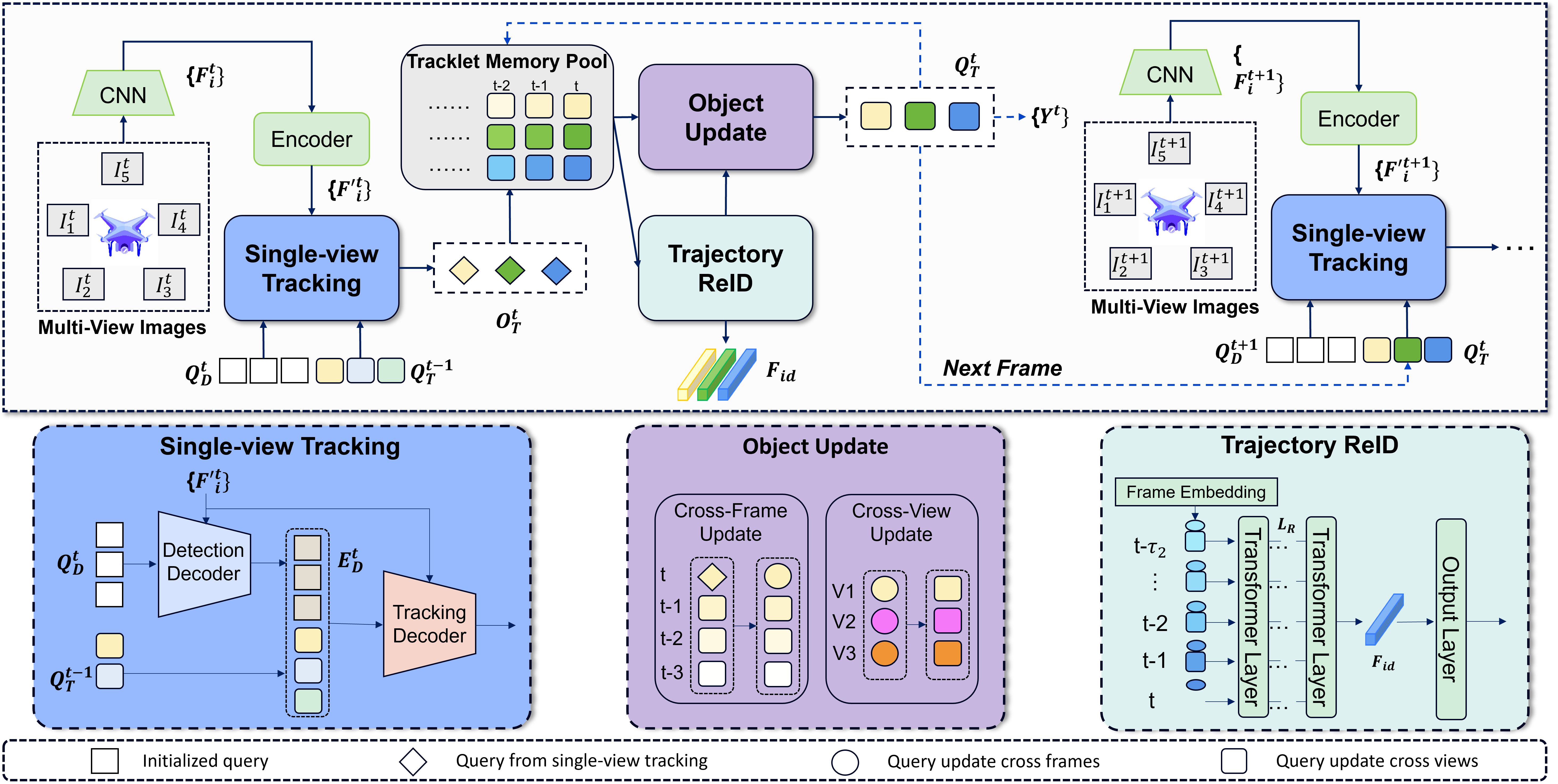}
		\caption{Overview of our FusionTrack framework. It comprises the Single-view Tracking module, which tracks objects independently within each view. The Tracklet Memory Pool stores queries collected from multiple views and across temporal frames. The Trajectory ReID module extracts discriminative ReID features to facilitate robust cross-view association. And the Object Update Module refines current-frame queries by integrating spatiotemporal context, producing enhanced representations.}
		\label{framework}
	\end{figure*}
	
	\subsection{Overview}
	We present \textbf{FusionTrack}, a novel Transformer-based framework for MVMOT as shown in Fig. \ref{framework}. Our approach fundamentally differs from existing methods by unifying SVT and cross-view association within a cohesive end-to-end architecture, enabling simultaneous optimization of detection, tracking, and ReID tasks. During training, our carefully designed Object Update Module (OUM) performs cross-frame and cross-view dynamic feature rectification, while the Tracklet Memory Pool (TMP) maintains consistent identity representations across both overlapping and non-overlapping views. For inference, we introduce a View-guided Hierarchical Clustering (VHC) algorithm coupled with a Neighbor Filtering Mechanism (NFM) to achieve robust cross-view identity association.
	
	
	\subsection{Problem Definition}
	Consider a multi-view object tracking system with $C$ drones and their cameras. Let $\mathcal{I} = \{ I_c^t \}_{t=1}^T$ denote the video sequence captured by the $c$-th camera, where $I_c^t$ represents the image frame at time $t$. Our objective is to track and obtain the tracklet set $\mathcal{T} = {T_i^c}$ for $N_c$ targets in each view $c$, where $c\in[1,C]$ and $i\in[1,N_c]$. For the multi-view tracking task, the state $t_s^i$ of each tracklet $T_i^c$ at time step $s$ can be represented as:
	\begin{equation}
		t_s^i = (B_s^i, CLS_s^i, ID_s^i),
	\end{equation}
	where $B_s^i = (x, y, w, h)$ denotes the target's bounding box coordinates, $CLS_s^i$ indicates the object class label, and $ID_s^i$ represents the target identity.
	
	
	\subsection{Pipline of FusionTrack}
	Recent advances in Transformer-based end-to-end detection and tracking frameworks (e.g., TransTrack\cite{sun2020transtrack}, MOTR\cite{zeng2022motr}, MeMOTR\cite{gao2023memotr}) have demonstrated superior performance by employing detect queries and track queries for tracklet maintenance within powerful feature representations. Building upon this paradigm, we extend these approaches to MVMOT, enabling joint training of SVT and cross-view ReID in a unified end-to-end framework.
	
	
	\subsubsection{Single view tracking}
	Prior to addressing cross-view ReID, we first complete single-view feature extraction and obtain detection/tracking queries. Following DETR-family methodologies, ResNet50\cite{he2016deep} extracts initial image feature maps $\mathbf{F}_i^{t}$ , which are subsequently refined through a Transformer encoder to enhance spatial contextual information, yielding updated feature maps $\mathbf{F}_i^{'t}$.
	Inspired by MOTR\cite{zeng2022motr}, two independent decoders are employed for object detection and cross-frame tracking. The detection decoder initially employs image-to-query cross-attention to optimize detection queries. Subsequently, the tracking decoder strategically integrates refined detection queries with track queries through self-attention and cross-attention mechanisms to acquire target bounding boxes for persistent tracking.
	We now elaborate on each component in detail.
	
	\textbf{Detection Decoder:}
	As shown in Fig. \ref{framework}, the detection decoder takes detection queries $\mathbf{Q}_D^t \in \mathbb{R}^{N\times d_{model}}$ and the current frame's feature map $\mathbf{F}_i^{'t}$ as inputs, generating object detection embeddings $\mathbf{E}_D^t \in \mathbb{R}^{N\times d_{model}}$:
	\begin{equation}
		\mathbf{E}_D^t = \text{Decoder}(\mathbf{Q}_D^t, \mathbf{F}_i^{'t}).
	\end{equation}
	
	\textbf{Tracking Decoder:}
	The tracking decoder combines $\mathbf{E}_D^t$ with track queries $\mathbf{Q}_T^{t-1} \in \mathbb{R}^{M_{t-1}\times d_{model}}$ from the previous frame. The concatenated queries and feature map $\mathbf{F}_i^{'t}$ are processed to produce tracking outputs $\mathbf{O}_T^t \in \mathbb{R}^{M_t\times d_{model}}$:
	\begin{equation}
		\mathbf{O}_T^t = \text{Decoder}\big(\text{Concat}\{\mathbf{Q}_T^{t-1}, \mathbf{E}_D^t\}, \mathbf{F}_i^{'t}\big).
	\end{equation}
	
	\textbf{Tracklet Memory Pool:}
	To handle temporary object disappearances and reappearances in video sequences, we introduce a TMP that stores track queries $\mathbf{Q}_T^t$ over a time window $\tau_1$ :
	\begin{equation}
		\text{TMP} = \{\mathbf{Q}_T^{t-\tau_1}, \dots, \mathbf{Q}_T^t\}.
	\end{equation}
	Notably, all queries in TMP from $t-\tau_1$ to $t-1$ are feature rectified. The current frame's query will be updated and reinserted into TMP in subsequent steps.

	\subsubsection{Combine with ReID}
	Each camera view generates track queries that encode the target’s appearance and motion context. Inspired by TransReID~\cite{he2021transreid}, we design a fully query-driven multi-view ReID module, as shown in Fig. \ref{framework}. The module consists of $L_R$ stacked transformer layers and takes a sequence of queries from the TMP as input.
	
	Given a time window of $\tau_2$ frames, the input query sequence is denoted as $\{Q_T^{t-\tau_2}, \dots, Q_T^t\}$, where $\tau_2 \in [1, \tau_1]$, and $\tau_1$ is the maximum memory length. To incorporate temporal alignment across views, we assign a globally unique frame index to each image frame during data loading. This allows temporal information to be consistently embedded into each token, regardless of the camera perspective.
	
	Before feeding the query sequence into the transformer, we compute a Frame Embedding (FE) $ \in \mathbb{R}^{\tau_2 \times d} $ using a sinusoidal encoding scheme. The frame numbers are employed to inject temporal ordering information of video frames while guiding appearance alignment at corresponding timestamps:
	
	\begin{equation}
		FE_{(frame, 2i)} = \sin \left(\frac{frame}{10000^{\frac{2i}{d}}} \right).
	\end{equation}
	\begin{equation}
		FE_{(frame, 2i+1)} = \cos \left(\frac{frame}{10000^{\frac{2i}{d}}} \right).
	\end{equation}
	
	This embedding is element-wise added to the query tokens incorporate source information. The resulting sequence is then fed into the transformer encoder (Fig. \ref{framework}) to produce a fused ReID feature representation $F_{\mathrm{id}} \in \mathbb{R}^{D}$. Finally, a softmax classification head is used to predict the tracklet identity, then the ID loss is computed.
	
	\subsection{Joint Loss Function for Tracking and ReID}
	
	To facilitate joint learning of tracking and ReID, we adopt a multi-task loss formulation. The total loss consists of two components: tracking loss $\mathcal{L}_T$ and ReID loss $\mathcal{L}_R$. Inspired by FairMOT~\cite{zhang2021fairmot}, we introduce an uncertainty-aware weighting strategy to dynamically balance the two tasks:
	
	\begin{equation}
		\mathcal{L}_{total} = \frac{1}{2} \left( \frac{1}{e^{\omega_1}} \mathcal{L}_{T} + \frac{1}{e^{\omega_2}} \mathcal{L}_{R} + \omega_1 + \omega_2 \right)
	\end{equation}
	where $\omega_1$ and $\omega_2$ are learnable parameters that adaptively control the contribution of each task during training.
	
	The tracking loss $\mathcal{L}_T$ is composed of three terms:
	
	\begin{equation}
		\mathcal{L}_{T} = \lambda_{cls} \mathcal{L}_{cls} + \lambda_{reg} \mathcal{L}_{reg} + \lambda_{giou} \mathcal{L}_{giou}
	\end{equation}
	Here, $\mathcal{L}_{cls}$ is the focal loss for classification, $\mathcal{L}_{reg}$ is the L1 loss for bounding box regression, and $\mathcal{L}_{giou}$ denotes the Generalized IoU loss. The weights $\lambda_{cls}$, $\lambda_{reg}$, and $\lambda_{giou}$ are empirically determined .
	
	The ReID loss $\mathcal{L}_R$ includes identity classification loss and triplet loss:
	
	\begin{equation}
		\mathcal{L}_{R} = \mathcal{L}_{id} + \mathcal{L}_{trip}
	\end{equation}
	where $\mathcal{L}_{id}$ is the cross-entropy loss for identity prediction, and $\mathcal{L}_{trip}$ is the triplet loss that encourages discriminative embedding learning.
	This combined loss design enables effective synergy between tracking and ReID, ultimately enhancing both accuracy and consistency across views.

	\subsection{Object Update Module}
	
	Upon completing SVT and multi-view identity alignment, each object is assigned a unified identity within the TMP. To enrich the object query features with stronger spatiotemporal context, we introduce OUM as a feature rectification module that refines the current-frame object queries through two complementary mechanisms: (1) Object Update Cross Frames: The module integrates object queries from historical frames to infuse temporal continuity into the current features. (2) Object Update Cross Views: It further updates the features by aggregating queries of the same object identity from other viewpoints, allowing the representation to adapt to view-specific appearance variations and thereby improving robustness.
	This update strategy enables the preservation of both long- and short-term context, which significantly stabilizes object tracking over time. Furthermore, by leveraging diverse object appearances captured from different views, OUM improves the discriminative power of query features.
	
	\begin{figure}[h]
		\centering
		\includegraphics[width=3.4in]{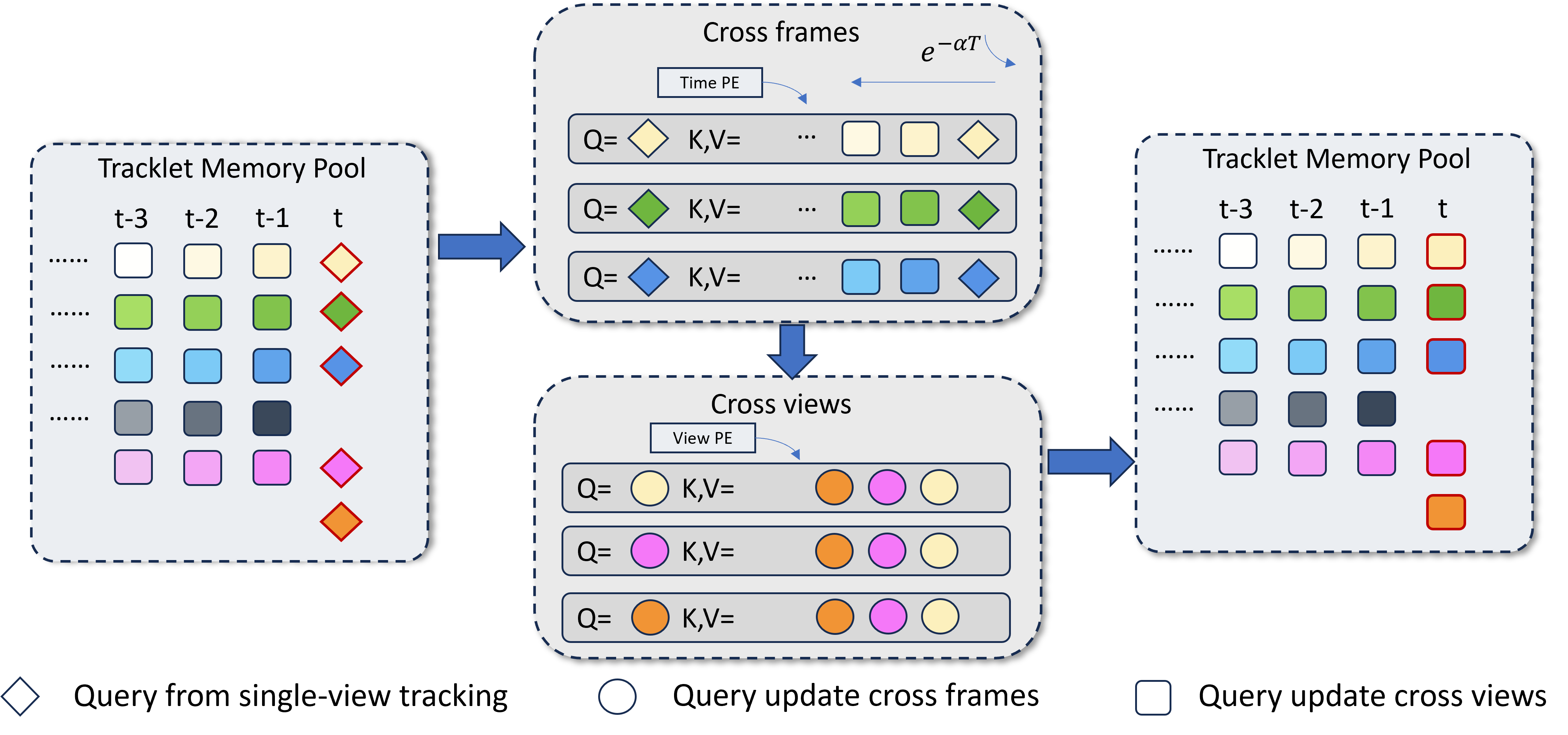}\caption{Pipline of the Object Update Module. We update the object queries cross frames and views.}
		\label{update}
	\end{figure}
	
	
	\subsubsection{Object Update Cross Frames}
	To enrich the current query representation with temporal context, we first perform temporal cross-attention over $\tau_3$ frames, as shown in Fig. \ref{update}. We assume that the TMP provides access to all object queries within the last $\tau_3$ frames. Since recent frames carry more relevant motion and appearance cues, we introduce a time-decayed attention mechanism, which assigns higher weights to queries closer to the current frame. Let $Q_T^t$ denote the object query at frame $t$, the updated query feature is computed as:
	
	
	
	\begin{equation}
		\begin{split}
			Q_T^{t} = \text{WeightedCrossFrameAttn}\big(
			&Q = Q_T^{t}, \\
			&K,V = \{Q_T^{t-\tau_3}:Q_T^{t}\}, \\
			&PE = \text{Pos}(t-\tau_3:t)
			\big)
		\end{split}
	\end{equation}
	The term $Pos(t{-}\tau_3{:}t)$ encodes temporal context into the positional representation. If some frames within the temporal window are missing, their features are excluded from attention computation. In essence, temporal decay is implemented by integrating a time-distance weight matrix $W$ into the cross-attention formulation. Let the current-frame object query serve as the query vector $Q$, and historical queries within the past $\tau_3$ frames serve as keys $K$ and values $V$. The modified attention score is computed as:
	
	\begin{equation}
		\text{score}= \text{Softmax}\left( \frac{Q K^\top}{\sqrt{d_k}} \odot W \right)
	\end{equation}
	where $Q \in \mathbb{R}^{1\times d_{model}}$, $K \in \mathbb{R}^{\tau_3\times d_{model}}$, $V \in \mathbb{R}^{\tau_3\times d_{model}}$, and $W = e^{-\alpha \cdot T} \in \mathbb{R}^{1\times \tau_3}$, with $T \in [0,\tau_3]$ representing the temporal decay weight.

	
	\subsubsection{Object Update Cross Views}
	A key challenge in multi-view collaborative tracking lies in the significant appearance variations of the same object across different viewpoints. To mitigate this, we leverage cross-view counterparts to refine object queries and reduce appearance discrepancies.
	As illustrated in Fig. \ref{update}, we employ a cross-view attention mechanism to aggregate contextual cues from different views, enabling more holistic representations of each object. Leveraging the ReID results obtained during training, we select identity-consistent objects from TMP module and use them as auxiliary references.
	Since ReID associations are unreliable at the beginning of training, cross-view updates are only activated after the model stabilizes. In Fig. \ref{update}, the blue, green, and yellow tokens represent query features of the same object from different viewpoints. The update is performed via cross-attention as follows:
	
	\begin{equation}
		\begin{split}
			Q_T^{t} = \text{CrossViewAttn}\big(Q, K, V = Q_{identity}^t,\\
			PE = \text{Pos}(1:C)\big)
		\end{split}
	\end{equation}
	where $Q_{identity}^t$ represents the set of query features of the same target identity in the current frame, and $PE = \text{Pos}(1:C)$ incorporates view information into the positional embedding.
	Through the above two-stage refinement, the query features are successively updated with both temporal and cross-view information. Rectified under a unified spatio-temporal reference, these enhanced features significantly improve the accuracy and robustness of object tracking and cross-view association.

	\begin{figure*}[!t]
		\centering
		\includegraphics[width=7in]{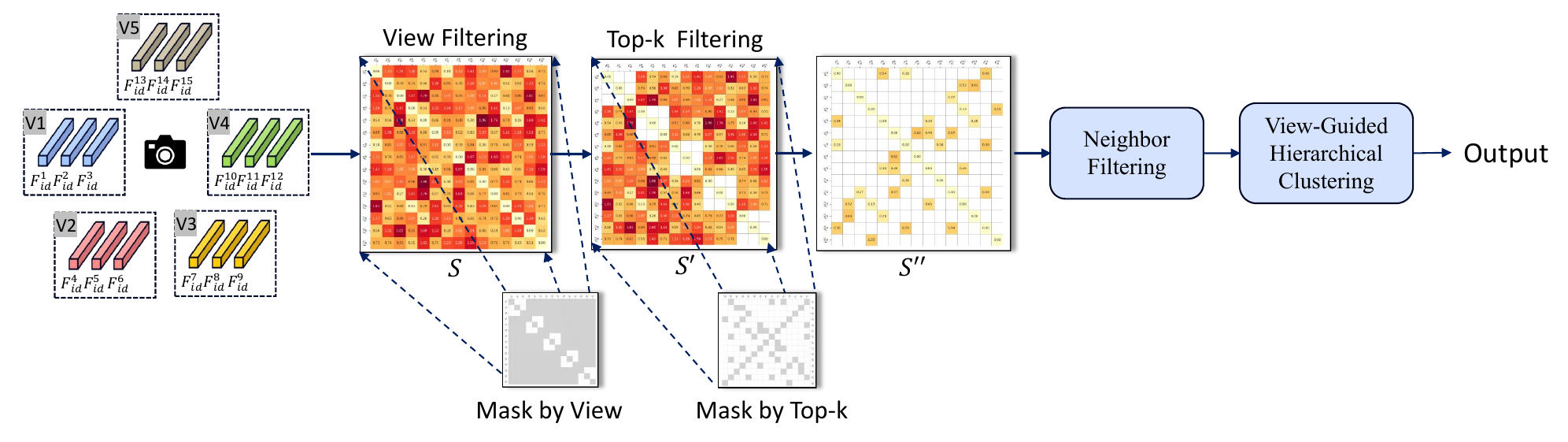}
		\caption{Post-processing pipeline, consisting of view-based masking, top-k masking, Neighbor Filtering, and Viewpoint-guided Hierarchical Clustering.}
		\label{postprocess}
	\end{figure*}
	\subsection{Inference Process}
	
	\subsubsection{Cross-view ReID}
	During inference, feature extraction follows the same design as in training. After obtaining tracking outputs and ReID features, we perform identity alignment across camera views.
	Specifically, we maintain a sliding window of output queries within the most recent $\tau_1$ frames, stored in the TMP. If an object is not detected in the current frame but has appeared recently, it is marked as \textit{inactive}. Should it remain undetected for more than $\tau_1$ frames, the corresponding query is discarded. If it reappears within the window, it is reactivated and relabeled as \textit{active}.
	For newly detected objects, the system creates new query entries in TMP to initialize their trajectories. The ReID module then extracts identity-aware features for each current-frame object using its query representation.
	While certain objects may lack a complete $\tau_2$-frame query sequence, our model can robustly process such cases due to the transformer's sequence modeling capability and the presence of frame-level positional embedding.
	
	During inference, cross-frame and cross-view feature updates remain necessary. However, since ground-truth identity labels are unavailable at test time, it is not possible to directly associate objects across views.
	To approximate identity correspondence, we compute pairwise cosine similarities between ReID features and apply a confidence threshold $\delta$ to determine potential cross-view matches. This strategy allows the model to perform reliable feature updates, ensuring consistent identity propagation in the absence of ground-truth supervision.
	
	\subsubsection{Neighbor Filtering Mechanism and Hierarchical Clustering}
	To assign object identities during inference, we introduce a VHC enhanced with a NFM. This design ensures that identity association respects both intra-view constraints and inter-view consistency.
	We define two key rules for similarity-based cross-view clustering:\\
	\textbf{Rule 1:} The distance between different objects from the same view is set to infinity, thereby avoiding intra-view associations.\\
	\textbf{Rule 2:} Once a tracklet $A$ from one view is associated with a tracklet $B$ from another view, the distance between $A$ and all other trajectories from the same view as $B$ is also set to infinity.

	As illustrated in Fig. \ref{postprocess}, we construct a pairwise distance matrix $S$ by computing cosine similarity between all object queries in the current frame. The matrix quantifies the appearance-based similarity across views and serves as the foundation for identity clustering. The formulation is given by:
	\begin{equation}
		S =\begin{bmatrix}
			\cos(F_{id}^1, F_{id}^1) & \cdots & \cos(F_{id}^1, F_{id}^N) \\
			\vdots & \ddots & \vdots \\
			\cos(F_{id}^N, F_{id}^1) & \cdots & \cos(F_{id}^N, F_{id}^N)
		\end{bmatrix}_{N_t \times N_t}
	\end{equation}
	Let $\cos$ denote cosine distance, $F_{id}^i$ the ReID feature of the $i$-th object, and $N_t$ the total number of objects in the current frame.
	
	Following Rule 1 we generate an initial mask matrix $M_1$ where intra-view pairs are assigned a value of $+\infty$ and inter-view pairs are set to 1. The masked distance matrix is then computed as $S' = S \odot M_1$, which ensures that identity assignments do not occur within the same view.
	To further refine potential identity matches, we adopt a NFM. For each object, we select its top-$k$ most similar candidates from other views based on $S'$, and retain only mutual top-$k$ pairs to construct the $k$-neighbor mask matrix $S''$.
	If an object does not appear in $S''$ as a valid neighbor of any cross-view object, it is considered visible only within its own view.
	Given the presence of numerous visually similar objects (e.g., vehicles) in complex urban scenes, relying solely on top-$k$ similarity may lead to identity confusion. To address this, we integrate a neighbor consistency check to eliminate spurious associations.
	
	
	
	
	The NFM leverages the spatial consistency typically exhibited among nearby objects under overlapping views. It imposes a critical constraint: if two objects are considered a valid match in $S''$, a significant proportion of their respective spatial neighbors should also mutually correspond.
	To quantify this constraint, we define a threshold $\delta'$: if more than $\delta'$ of the neighbors of a matched object pair are also found within each other’s top-$k$ neighbor lists, the match is retained; otherwise, it is filtered out. This mechanism ensures that the retained distance pairs exhibit both appearance similarity and spatial context consistency.
	
\begin{figure}[h]
		\centering
		\includegraphics[width=3.4in]{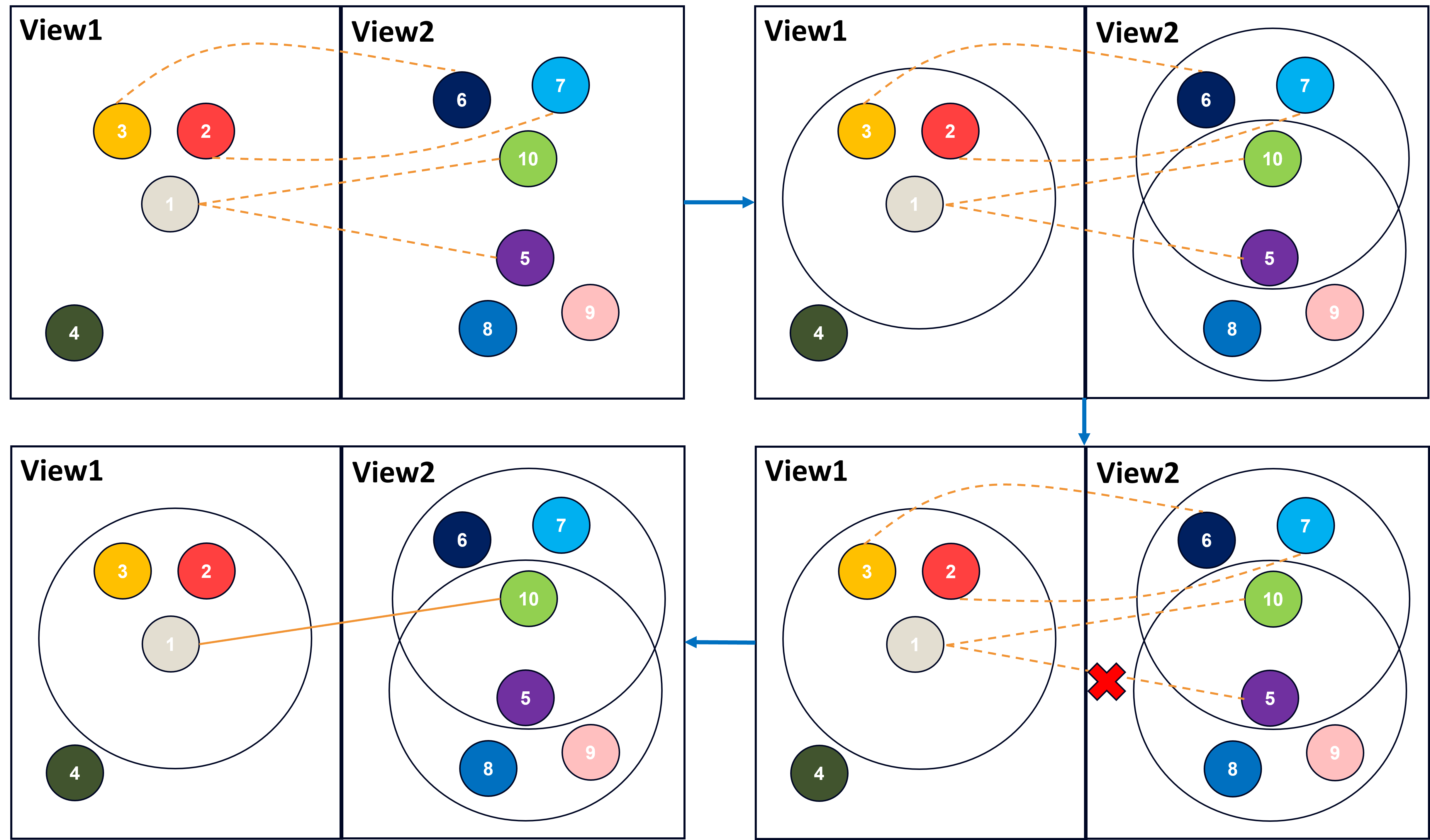}
		\caption{Implementation of the NFM. Dashed lines indicate mutual top-k relationships between objects.}
		\label{neighbor}
	\end{figure}

	As illustrated in Fig. \ref{neighbor}, consider an example where object $\sharp$1 in the left view is mutually matched with objects $\sharp$5 and $\sharp$10 in the right view. We examine whether the spatial neighbors of object $\sharp$1 (e.g., objects $\sharp$2 and $\sharp$3) are connected to neighbors of object $\sharp$5 and $\sharp$10 (e.g., objects $\sharp$6 and $\sharp$7). If most of these secondary neighbors appear in the corresponding top-$k$ lists, the association is considered valid. Otherwise, mismatched associations (e.g., involving objects $\sharp$8 and $\sharp$9) are discarded.

	
	Finally, we conduct cross-view identity association using VHC. Specifically, after each clustering operation, we apply Rule 2 to modify the distance matrix, preventing any object from being matched with multiple candidates across different views.
	Once the clustering process reaches a predefined termination criterion, we obtain the final set of cross-view identity associations	.
	
	Through the proposed inference pipeline, we achieve reliable multi-view multi-object trajectories. Experimental results validate the robustness and effectiveness of our framework.
	
	\subsection{Motivation Analysis}
	
	In a sense, our framework is grounded in a multi-task learning paradigm, where the original SVT objective is extended to a parallel cross-view ReID task. This design enables the model to extract consistent and discriminative representations for the same identity across different viewpoints using a unified object query encoder.
	Leveraging contrastive loss, the model effectively learns appearance variations of identical targets while discriminating different targets with similar visual characteristics (e.g., color, texture), thereby enhancing the robustness of feature extraction in tracking processes. Moreover, the filtering mechanism introduced during inference further enhances the robustness of the ReID matching process.
	
	\section{EXPERIMENTS}

	\subsection{Implementation details}
	The proposed FusionTrack employs ResNet50\cite{he2016deep} as the backbone network, initialized with DAB-Deformable-DETR\cite{liu2201dab} weights pre-trained on the COCO\cite{lin2014microsoft} dataset. For the MDMOT dataset, input images are resized to $1280\times720$ resolution while maintaining aspect ratio, with data augmentation techniques including random scaling and cropping. All experiments are conducted on two NVIDIA A800 GPUs using PyTorch, with a batch size of 1 (each containing multi-view video sequences) and the Adam optimizer (initial learning rate: $1\times10^{-5}$) for 80 epochs. To address cross-view alignment instability during early training stages, we adopt a progressive strategy: we first disable cross-view matching for the first 20 epochs. Then, we optimize the ReID feature extraction structure after query features stabilize. Finally, we refine queries simultaneously using aligned views.
	
	In the detection decoder module, we initialize 300 detection queries ($Q^t_D$), consistent with MOTR\cite{zeng2022motr}. The trajectory memory pool retains $\tau_1$=30 frames, while the tracklet sequences fed into the ReID module and the cross-frame update queries contain $\tau_2$=10 and $\tau_3$=6 frames, respectively. A temporal decay factor $\alpha$ of 0.5 is applied to emphasize current frame features. Following FairMOT\cite{zhang2021fairmot}, the loss balancing parameters $w_1$ and $w_2$ are initialized as -1.85 and -1.05.
	
	During inference, a post-processing module enhances ReID features for cross-view association. Initial matching requires a confidence threshold $\delta$ of 0.8 to prioritize high-similarity targets for subsequent cross-view feature updates and neighbor filtering. For mutual K-nearest neighbor reasoning, candidate targets must be mutually ranked within the top-10 similarity scores. The NFM employs a threshold $\delta'$ of 0.5 to eliminate unreliable associations. All other parameters remain identical to training configurations.
	\subsection{Evaluation metrics}
	The primary distinction between MVMOT and SVT lies in the ability to associate the same target across different views under a unified identity. To evaluate our approach, we adopt the CVMA and CVIDF1 metrics proposed in \cite{hao2024divotrack}, which extend the SVT metrics MOTA and IDF1 to the multi-view setting. The formulas for these metrics are defined as follows:
	
	\begin{equation}
		CVMA = 1 - \left( \frac{\sum_t m_t + fp_t + 2mme_t}{\sum_t gt_t} \right)
	\end{equation}
	
	\begin{equation}
		CVIDF1 = \frac{2 \times CVIDP \times CVIDR}{CVIDP + CVIDR}
	\end{equation}
	where $m_t$, $fp_t$, $mme_t$, and $gt_t$ represent the number of \textit{misses}, \textit{false positives}, \textit{mismatched pairs}, and the total number of tracked targets across all views at time $t$, respectively. Additionally, CVIDP and CVIDR denote the cross-view identity precision and recall, respectively.
	
	These metrics effectively measure the accuracy and robustness of multi-view identity association, providing a comprehensive evaluation of MVMOT performance.
	\begin{table*}[!t]
		\caption{COMPARISON OF SVT METRICS ON MDMOT DATASET. THE BEST AND SECOND-BEST PERFORMANCES AMONG THE TRANSFORMER-BASED METHODS ARE SHOWN IN BOLD AND LIGHT BLUE.}
		\centering
		\begin{tabular}{l|ccccccccc}
			\toprule
			Methods&MOTA↑&MOTP↑&IDF1↑&MT↑&ML↑&HOTA↑&FP↓&FN↓&IDS↑ \\
			
			\midrule
			\textit{CNN-based}:\\
			DeepSORT\cite{wojke2017simple} (2017) & 84.12 & 84.45 & 88.32 & 520 &50 & 79.7 & 45324 & 28532 & 280 \\
			Centertrack\cite{zhou2020tracking} (2020) & 88.36 & 87.60 & 92.41 & 740&29  & 83.33 & 37698 & 21913 & 248 \\
			FairMOT\cite{zhang2021fairmot} (2021) & 88.96 & 87.32 & 91.8 & 709&27 & 86.3 & 37072 & 19876 & 103 \\
			TraDes\cite{wu2021track} (2021) & 86.78 & 86.66 & 91.38 & 733&41 & 81.4 & 43589 & 24189 & 221 \\
			Bytetrack\cite{zhang2022bytetrack} (2022) & 86.38 & 87.94 & 91.96 & 703&48 & 86.9 & 34506 & 29468 & 135 \\
			OC-SORT\cite{cao2023observation} (2023) & 89.42 & 88.73 & 92.15 & 768&18 & 85.12 & 32105 & 18542 & 102 \\
			\midrule
			\textit{Transformer-based}:\\
			TransTrack\cite{sun2020transtrack} (2020) & 84.17 & 86.42 & 89.37 & 498&47 & 82.35 & 35542 & 32584 & 202 \\
			MOTR\cite{zeng2022motr} (2022) & 87.03 & 86.82 & 91.13 & 634&57 & 81.42 & 38945 & 27102 & 103 \\
			MeMOTR\cite{gao2023memotr} (2023) & \textcolor{blue}{87.57} & \textcolor{blue}{87.69} & \textcolor{blue}{91.97} & \textcolor{blue}{719}&\textcolor{blue}{21} & \textcolor{blue}{83.94} & \textcolor{blue}{31984} & \textcolor{blue}{24980} & \textcolor{blue}{92} \\
			
			\midrule
			\rowcolor{gray!20}
			\textbf{Ours} & \textbf{88.13} &\textbf{87.92} & \textbf{92.04} & \textbf{745}&\textbf{19} &\textbf{ 84.42} & \textbf{30876} & \textbf{23954} & \textbf{83}\\
			
			\bottomrule
		\end{tabular}
		\label{table2}
	\end{table*}
	\begin{table*}[!t]
		\caption{COMPARISON OF MVMOT RESULTS ON THE EPFL, CAMPUS, MVMHAT, WILDTRACK AND DIVOTRACK. THE BEST AND SECOND-BEST PERFORMANCES FOR EACH COLUMN ARE SHOWN IN BOLD AND LIGHT BLUE. }
		\centering
		\begin{tabular}{l|cccccccccc}
			\toprule
			& \multicolumn{2}{c}{CAMPUS}& \multicolumn{2}{c}{WILDTRACK} & \multicolumn{2}{c}{MvMHAT}& \multicolumn{2}{c}{DIVOTrack}\\

			\midrule
			Methods&CVMA↑&CVIDF1↑&CVMA↑&CVIDF1↑&CVMA↑&CVIDF1↑&CVMA↑&CVIDF1↑ \\
			\midrule
			OSNet\cite{zhou2019omni}(2019)&58.8&47.8&10.8&18.2&\textcolor{blue}{92.6}&\textcolor{blue}{87.7}&33&44.9 \\
			Strong\cite{luo2019bag}(2019)&63.4&55&28.6&41.6&49&55.1&39.1&44.7  \\
			AGW\cite{ye2021deep}(2021)&60.8&52.8&15.6&23.8&92.5&86.6&15.6&23.8 \\
			CT\cite{wieczorek2021unreasonable}(2021)&63.7&55&19&42&46.7&53.5&64.9&65 \\
			MGN\cite{wang2018learning}(2020)&63.3&56.1&32.6&46.2&92.3&87.4&33.5&39.4  \\
			MvMHAT\cite{gan2021self}(2021)&56&55.6&10.3&16.2&70.1&68.4&61.1&62.6 \\
			CrossMOT\cite{hao2024divotrack}(2024)&\textcolor{blue}{65.6}&\textcolor{blue}{61.2}&\textcolor{blue}{42.3}&\textcolor{blue}{56.7}&92.3&87.4&\textcolor{blue}{72.4}&\textcolor{blue}{71.1}  \\
			\midrule
			\rowcolor{gray!20}
			\textbf{Ours} & \textbf{68.8} & \textbf{62.5} & \textbf{53.5} & \textbf{60.8} & \textbf{92.8} & \textbf{88.2} & \textbf{74.5} & \textbf{77.3} \\
			
			\bottomrule
		\end{tabular}
		\label{table3}
	\end{table*}
	\subsection{Comparison with other SOTA methods}
	
	\subsubsection{Single-view tracking results comparison in MDMOT}
	
	To fully evaluate the MDMOT dataset, we first tested a range of widely adopted SVT methods, as shown in Table \ref{table2}. The comparison includes CNN-based methods such as DeepSORT\cite{wojke2017simple}, CenterTrack\cite{zhou2020tracking}, FairMOT\cite{zhang2021fairmot}, TraDes\cite{wu2021track}, ByteTrack\cite{zhang2022bytetrack}, and OC-SORT\cite{cao2023observation}, as well as Transformer-based methods like TransTrack\cite{sun2020transtrack}, MOTR\cite{zeng2022motr}, and MeMOTR\cite{gao2023memotr}. All methods were trained and tested with their official default configurations. 
	Among the Transformer-based approaches, MeMOTR achieves outstanding performance with MOTA of 87.57\%, IDF1 of 91.97\%, and HOTA of 83.94\%. The FusionTrack framework proposed in this study also adopts the Transformer-based architecture for SVT. It surpasses most CNN-based methods and achieves the best performance among Transformer methods, exceeding the second-best method, MeMOTR, by 0.56\% in MOTA and 0.23\% in MOTP.
	
	Our proposed FusionTrack maintains the single-view processing pipeline while introducing novel mechanisms, consequently achieving state-of-the-art results among Transformer-based methods. This improvement stems from three key designs: (a) joint optimization of ReID and SVT, (b) a multi-task loss function, and (c) cross-frame and cross-view feature updating strategies during both training and inference.These innovations significantly enhance prediction and association accuracy across diverse viewpoints, leading to superior SVT performance.

	\subsubsection{Multi-view tracking results comparison in pedestrian datasets}

	
	
	To evaluate the effectiveness of the proposed FusionTrack method in multi-view tracking, we tested it on four widely used multi-view multi-object pedestrian tracking datasets: CAMPUS\cite{xu2016multi}, WILDTRACK\cite{chavdarova2017wildtrack}, MvMHAT\cite{gan2021self}, and DIVOTrack\cite{hao2024divotrack}. The comparison methods included OSNet\cite{zhou2019omni}, Strong\cite{luo2019bag}, AGW\cite{ye2021deep}, CT\cite{wieczorek2021unreasonable}, MGN\cite{wang2018learning}, MvMHAT\cite{gan2021self}, and CrossMOT\cite{hao2024divotrack}. As shown in Table \ref{table3}, our method achieved outstanding results.
	
	On the CAMPUS dataset, our method reached 66.8\% in CVMA and 62.5\% in CVIDF1, outperforming all other methods. Compared to the second-best method, CrossMOT, our method improved CVMA by 1.2\% and CVIDF1 by 1.3\%. On the WILDTRACK dataset, despite fewer scenes and noisy annotations, our method still achieved 53.5\% and 58.0\%, surpassing CrossMOT by 11.2\% and 4.1\%, respectively, demonstrating superior robustness. On the MvMHAT dataset, our method achieved 92.8\% in CVMA and 88.2\% in CVIDF1, slightly surpassing OSNet and achieving the best overall performance. On the DIVOTrack dataset, our method achieved 74.5\% in CVMA and 77.3\% in CVIDF1, outperforming CrossMOT by 2.3\% and 6.1\%, showing superior performance in dense multi-object tracking scenarios.
	
	Overall, our method outperformed existing advanced methods in all evaluation metrics. This superiority is attributed to several factors: (a) jointly optimizing ReID and SVT allows the two tasks to mutually benefit from each other; (b) the feature update implemented through the OUM strengthens the model's robustness by effectively addressing challenges such as variations in viewpoint, lighting, and noise; (c) the Transformer architecture’s long-range dependency modeling ability enables the model to capture global target features more effectively, improving the accuracy of cross-view matching.
	
	\subsubsection{Multi-view tracking results comparison in MDMOT}
	\begin{table}[!h]
		\caption{COMPARISON OF MVMOT RESULTS ON THE MDMOT DATASET. THE BEST AND SECOND-BEST PERFORMANCES FOR EACH COLUMN ARE SHOWN IN BOLD AND LIGHT BLUE.}
		\centering
		\begin{tabular}{l|cc}
			\toprule
			&\multicolumn{2}{c}{MDMOT}\\
			\midrule
			Methods&CVMA&CVIDF1 \\
			\midrule
			OSNet\cite{zhou2019omni}(2019)&50.1&45.5\\
			Strong\cite{luo2019bag}(2019)&52.3&51.4\\
			AGW\cite{ye2021deep}(2021)&49.3&49.9\\
			Citytrack\cite{yang2022box}(2021)&57.27&56.09\\
			MvMHAT\cite{gan2021self}(2021)&\textcolor{blue}{78.83}&63.72 \\
			CrossMOT\cite{hao2024divotrack}(2024)&78.47&\textcolor{blue}{72.85}  \\	
			\midrule
			\rowcolor{gray!20}
			\textbf{Ours}&\textbf{80.8}&\textbf{75.2}  \\		
			\bottomrule
			
		\end{tabular}
		\label{table4}
	\end{table}
	To assess the applicability of the proposed dataset for multi-view tracking tasks and the effectiveness of the FusionTrack algorithm on multi-view UAV datasets, we performed experiments comparing FusionTrack with methods such as OSNet\cite{zhou2019omni}, Strong\cite{luo2019bag}, AGW\cite{ye2021deep}, CityTrack\cite{yang2022box}, MvMHAT\cite{gan2021self}, and CrossMOT\cite{hao2024divotrack}. The results, shown in Table \ref{table4}, demonstrate that FusionTrack achieves 80.79\% in CVMA and 75.21\% in CVIDF1, outperforming the best existing methods, MvMHAT and CrossMOT, by 2.0\% and 2.3\%, respectively.
	
	Compared with previous multi-view pedestrian datasets, our MDMOT dataset features top-down perspectives, mobility, and wide-area coverage. Despite these challenges, our method still achieves competitive performance, thanks to the collaborative design of each module, demonstrating the generalizability of our approach.
	\begin{table*}[!t]
		
		\caption{ABLAITION STUDY RESULTS OF EACH MODULE. }
		\centering
		\begin{tabular}{l|cccc|cc}
			\toprule
			&TMP&OUM(cross frames)&OUM(cross views)&NFM&CVMA&CVIDF1\\
			\midrule
			
			Base model&&&&&38.4&23.1\\
			Base model+TMP&\checkmark&&&&77.6&71.5\\
			Base model+TMP+OUM(cross frames)&\checkmark&\checkmark&&&79.4&73.1\\
			Base model+TMP+OUM&\checkmark&\checkmark&\checkmark&&80.6&74.8 \\
			\rowcolor{gray!20}
			\textbf{FusionTrack}&\checkmark&\checkmark&\checkmark&\checkmark&\textbf{80.8}&\textbf{75.2} \\

			\bottomrule
		\end{tabular}
		\label{ablation}
	\end{table*}
	\subsection{Ablation studies}
	
	In this section, we present ablation studies to evaluate the impact of each component of FusionTrack on performance, focusing on the TMP, OUM, and NFM. The OUM is divided into OUM (cross frames) and OUM (cross views). The experiments are carried out using the proposed MDMOT dataset. The base model performs basic SVT and ReID tasks collaboratively. However, as ReID features are computed considering only a single frame, identity mismatches occur frequently during cross-view association, resulting in poor performance across both metrics, with CVMA and CVIDF1 of 38.4 and 23.1 respectively.  As shown in the Table \ref{ablation}, each module contributes significantly to improving the final performance. In particular, after incorporating TMP, identity switches were significantly reduced, resulting in improvements to both CVMA and CVIDF1, which reached 77.6 and 71.5. Furthermore, OUM (cross frames) improves CVMA and CVIDF1 by 1.8\% and 2.4\%, respectively, by considering feature correlations across frames, ensuring feature propagation between frames. The introduction of OUM (cross views) further enhances the performance by 1.2\% and 0.9\%, effectively addressing the appearance variations of the same target across different viewpoints. The NFM mechanism eliminates unnecessary potential associations, improving performance by 0.2\% and 0.4\% in both metrics, further confirming the effectiveness of our approach. 
	
	The TMP is critical for our method, as it not only ensures continuous tracking of occluded targets during SVT but also provides a larger temporal span for feature extraction, resulting in more robust ReID features. The experimental results demonstrate that TMP is essential in our model, as its absence leads to frequent identity switches between frames, negatively impacting the performance.

	\subsection{Parameters selection}
	
	\begin{figure}[h]
		\centering
		\includegraphics[width=3.4in]{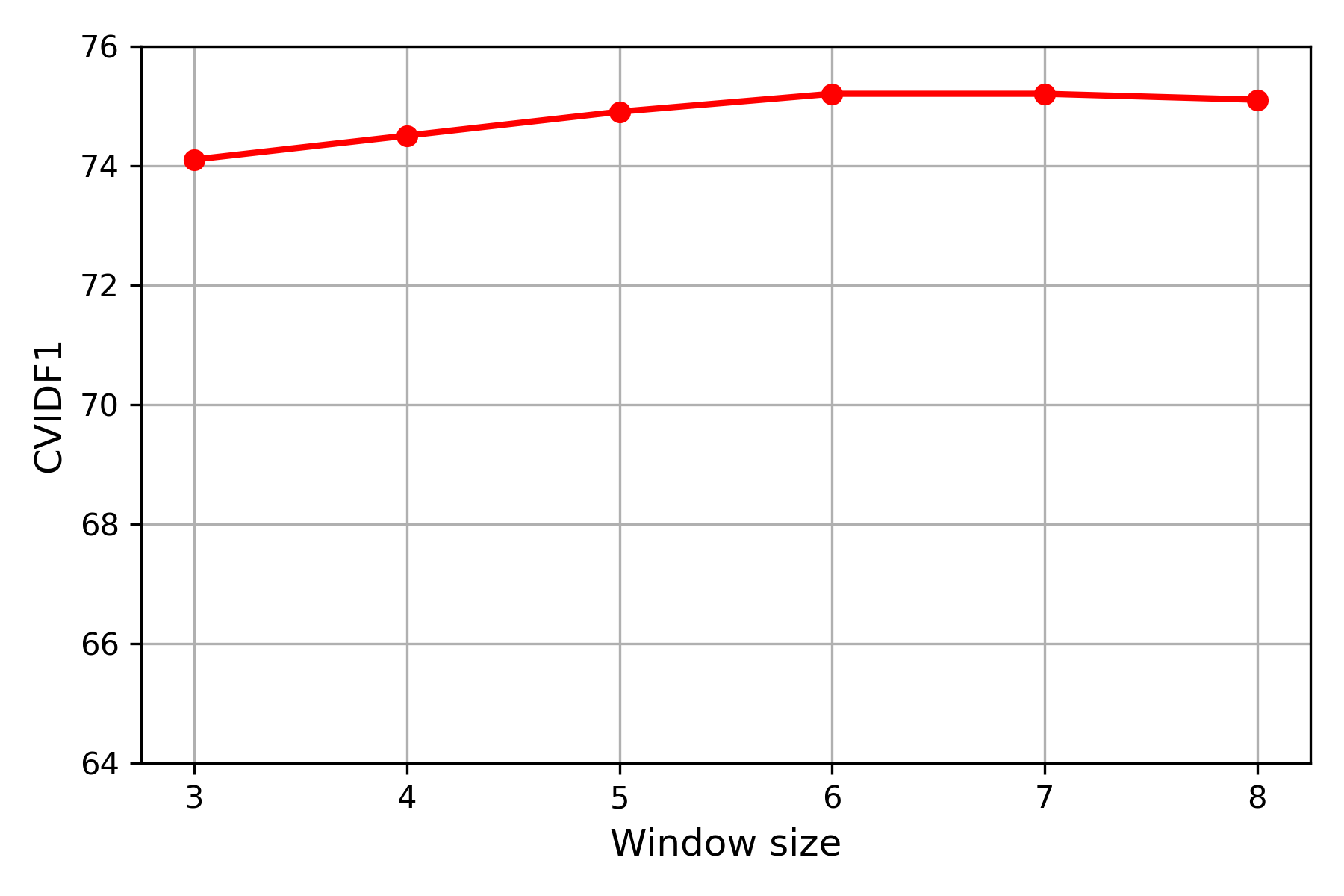}
		\caption{Variation of CVIDF1 with the change in cross-frame update window size.}
		\label{windows}
	\end{figure}
	In the experiments, two parameter settings significantly influenced the results. The first is the size of the historical frame window used in the cross-frame update of object features. We tested window sizes ranging from 3 to 8 frames, and the results, shown in the Fig. \ref{windows}, demonstrate that the 6-frame window size is the optimal choice, offering a good balance between accuracy and efficiency.
	
	\begin{figure}[h]
		\centering
		\includegraphics[width=3.4in]{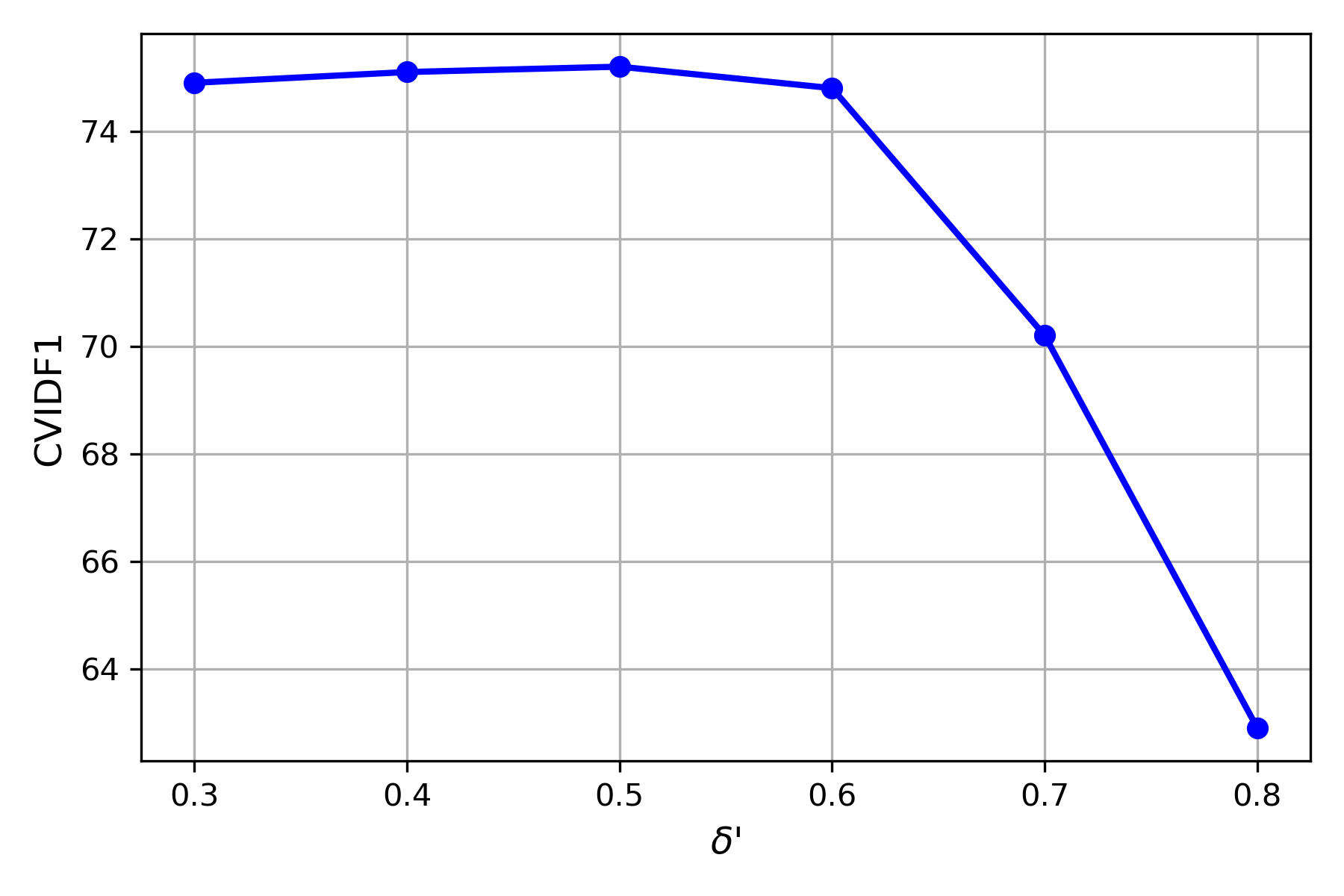}
		\caption{Variation of CVIDF1 with the change in NFM threshold $\delta$.}
		\label{delta}
	\end{figure}
	Furthermore, during inference, our NFM considers the ratio threshold of the potential neighbors' distances in the top-k set. A ratio that is too low will fail to eliminate irrelevant targets, while a ratio that is too high may exclude targets that should be associated, negatively impacting the overall target association results. As illustrated in the Fig. \ref{delta}, setting the threshold too low effectively eliminates filtering, leading to a slight performance drop. On the other hand, setting it too high excludes valid target associations, causing a significant performance decrease. After evaluation, we selecte 0.5 as the optimal threshold for NFM.

	\begin{figure*}[!t]
		\centering
		\includegraphics[width=6in]{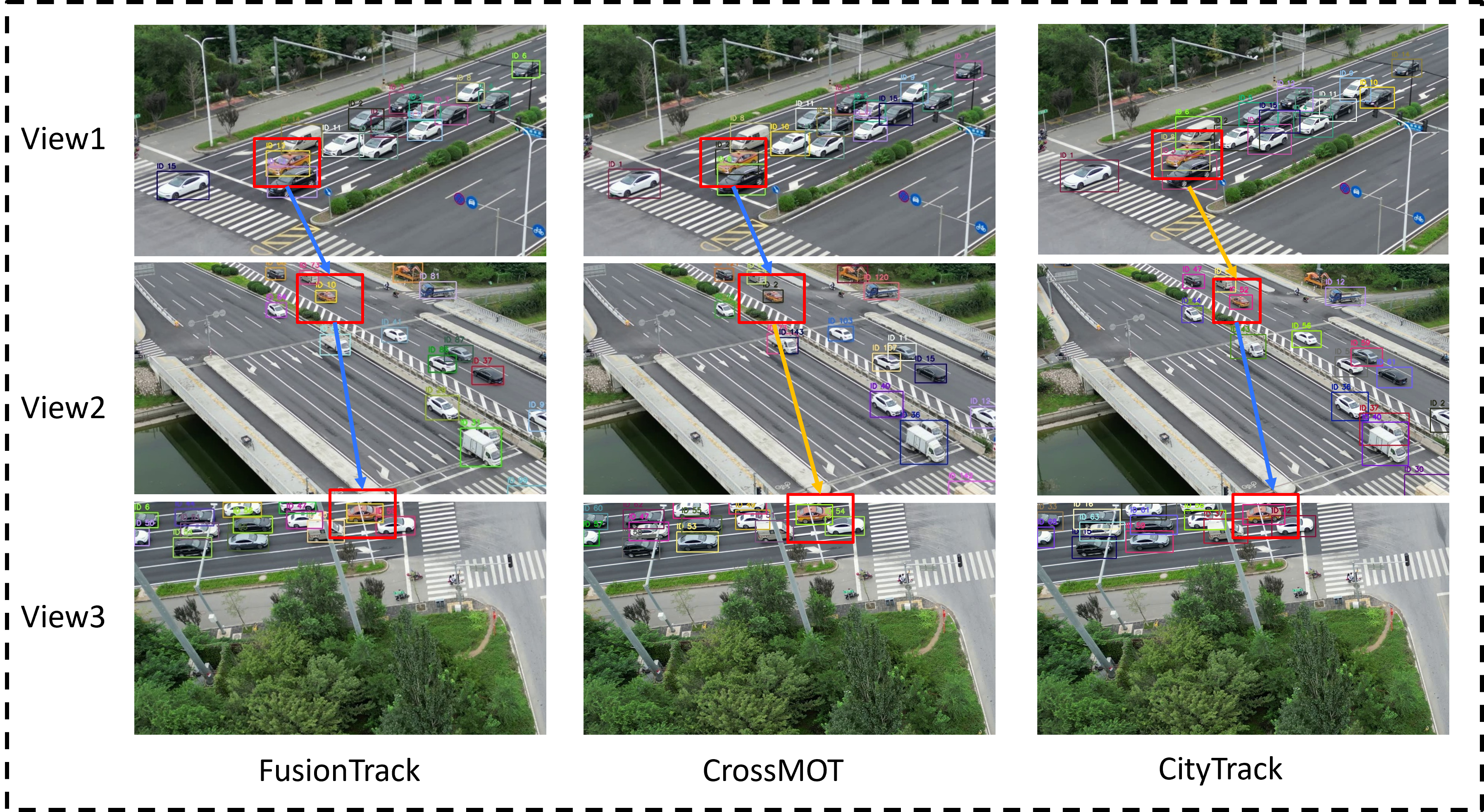}
		\caption{Illustrative results of associated tracking in intersection scene.}
		\label{vis1}
	\end{figure*}

	\begin{figure*}[!t]
		\centering
		\includegraphics[width=6in]{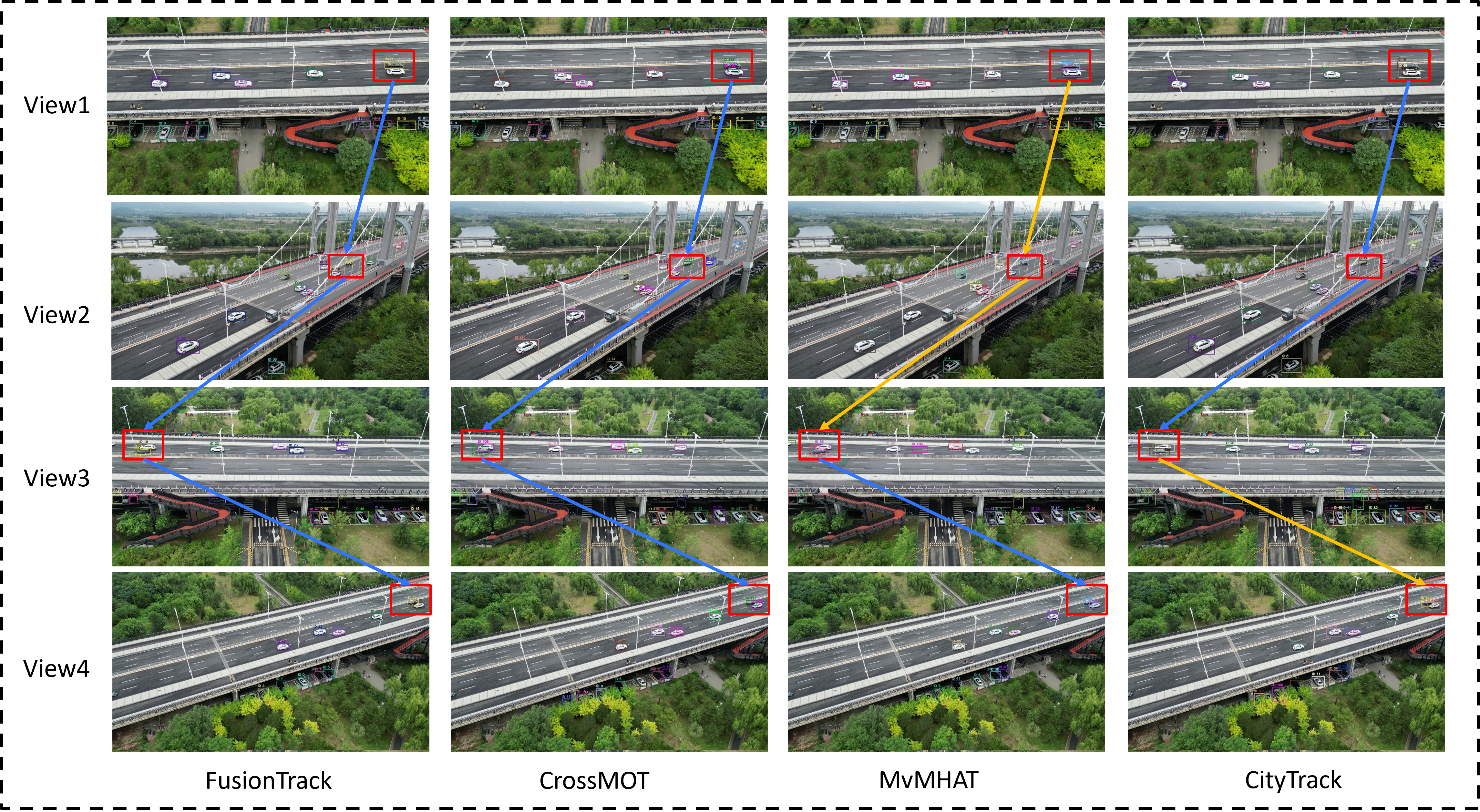}
		\caption{Illustrative results of associated tracking in overpass scene.}
		\label{vis2}
	\end{figure*}

	\subsection{Qualitative results}
	To provide a more intuitive comparison of method performance on our dataset, we conducted a qualitative analysis by visualizing the results from FusionTrack, CrossMOT\cite{hao2024divotrack}, MvMHAT\cite{gan2021self}, and CityTrack\cite{yang2022box}. As shown in Fig. \ref{vis1} and Fig. \ref{vis2}, the rows and columns represent different views and methods, respectively, with blue and yellow arrows indicating correct and incorrect matches. In Fig. \ref{vis1}, among the three views, View1 and View3 overlap partially, while View2 has no overlap with either View1 or View3. Fig. \ref{vis2} shows four views, with View1, View3 and View4 having partial overlap, while View2 is almost independent of the other three. The visualization results clearly show that our method significantly reduces cross-view target association errors compared to the other methods and performs best in multi-view tracking tasks. This advantage can be attributed to the integration of OUM and NFM in our framework, which together enhance the model's ability to effectively track objects across views.
	
	\section{CONCLUSION}
	We propose FusionTrack, an end-to-end Transformer-based framework for multi-view multi-object tracking that effectively integrates single-view object tracking and cross-view object association. The proposed architecture incorporates three key components: a Tracklet Memory Pool (TMP) for temporal feature propagation, an Object Update Module (OUM) for feature representation enhancement, and a Neighbor Filtering Mechanism (NFM) for efficient cross-view association. We additionally construct MDMOT, the first large-scale drone-view tracking benchmark supporting both overlapping and non-overlapping view. Our method demonstrates state-of-the-art performance on MDMOT and existing benchmarks. Comprehensive experiments and ablation studies confirm the approach's effectiveness and generalization capability, establishing a robust foundation for complex multi-view tracking applications.
	
	
	
	
	\bibliography{ref2}

	\bibliographystyle{IEEEtran}
	\newpage
	
	\vspace{11pt}
	
	\begin{IEEEbiography}[{\includegraphics[width=1in,height=1.25in,clip,keepaspectratio]{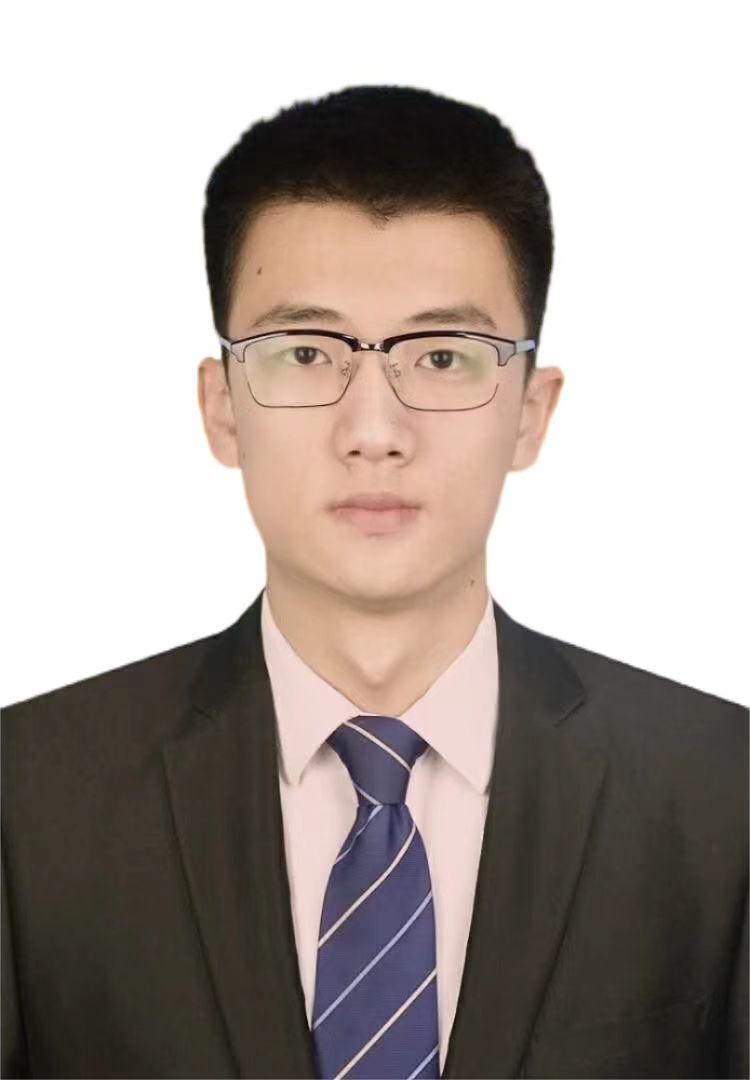}}]{Xiaohe Li}
	received the M.Sc degree from School of Software, Tsinghua University in 2021. He is currently an assistant research fellow at Aerospace Information Research Institute, Chinese Academy of Sciences. His research interests are focused on graph neural network, information fusion and collaborative awareness.
	\end{IEEEbiography}
	
	\begin{IEEEbiography}[{\includegraphics[width=1in,height=1.25in,clip,keepaspectratio]{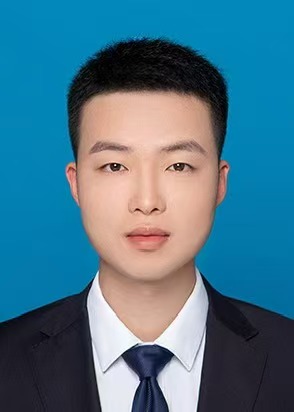}}]{Pengfei Li}
	is currently a master’s student with the Aerospace Information Research Institute, Chinese Academy of Sciences. His research interests include the multi-object detection and tracking.
	\end{IEEEbiography}
	\begin{IEEEbiography}[{\includegraphics[width=1in,height=1.25in,clip,keepaspectratio]{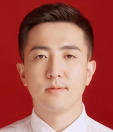}}]{Zide Fan (Member, IEEE)}
	received the Ph.D degree from Central South University in 2016. He is currently an associate research fellow at Aerospace Information Research Institute, Chinese Academy of Sciences. His research interests are focused on geographic data mining, information fusion.
	\end{IEEEbiography}
	
	\begin{IEEEbiography}[{\includegraphics[width=1in,height=1.25in,clip,keepaspectratio]{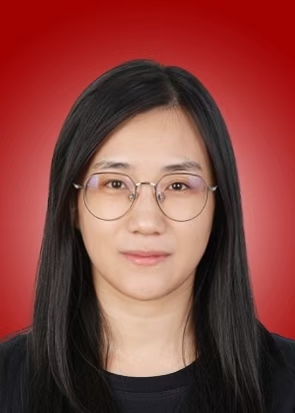}}]{Ying Geng}
	is currently an engineer at Aerospace Information Research Institute, Chinese Academy of Sciences. Her research interests are focused on aerospace simulation.
	\end{IEEEbiography}
	\begin{IEEEbiography}[{\includegraphics[width=1in,height=1.25in,clip,keepaspectratio]{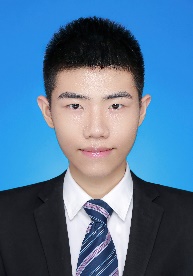}}]{Fangli Mou (Member, IEEE)}
	received the B.Sc. degree from Harbin Engineering University, Harbin, China, in 2015, and the M.Sc. and Ph.D. degrees from Tsinghua University, Beijing, in 2018 and 2023.
	He is an Assistant Professor with the Aerospace Information Research Institute, Chinese Academy of Sciences, Beijing. His research interests include computer vision, machine learning, robotics, and remote sensing processing.
	\end{IEEEbiography}
	\begin{IEEEbiography}[{\includegraphics[width=1in,height=1.25in,clip,keepaspectratio]{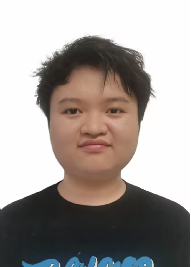}}]{Haohua Wu}
	received the B.Sc. degree from Beihang University, Beijing, China, in 2023.He is currently pursuing the M.Sc. degree with the University of Chinese Academy of Sciences, Beijing, China, and the Aerospace Information Research Institute, Chinese Academy of Sciences, Beijing.
	His research interests include distributed remote sensing image processing and deep learning.
	\end{IEEEbiography}
	
	\begin{IEEEbiography}[{\includegraphics[width=1in,height=1.25in,clip,keepaspectratio]{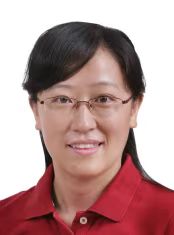}}]{Yunping Ge}
	is currently a senior engineer at Aerospace Information Research Institute, Chinese Academy of Sciences. Her research interests are focused on Network Information System.
	\end{IEEEbiography}
	

	\vfill
\end{document}